\documentclass[11pt]{article}

\usepackage[preprint]{acl}

\usepackage{times}
\usepackage{latexsym}

\usepackage[T1]{fontenc}

\usepackage[utf8]{inputenc}

\usepackage{microtype}

\usepackage{inconsolata}

\usepackage{graphicx}

\usepackage{booktabs}
\def\thickhline{\noalign{\hrule height 1.2pt}}
\usepackage{xcolor}
\usepackage{graphicx}
\usepackage[multiple]{footmisc}
\usepackage{orcidlink}

\title{Vectors from Larger Language Models Predict Human Reading Time and fMRI Data More Poorly when Dimensionality Expansion is Controlled}

\author{Yi-Chien Lin \\
  The Ohio State University \\
  \texttt{lin.4434@osu.edu} \\\And
  Hongao Zhu \\
  University of California \\ 
  San Diego \\
  \texttt{hoz034@ucsd.edu} \\\And
  William Schuler \\
  The Ohio State University \\
  \texttt{schuler.77@osu.edu} \\
  }

\begin{document}
\maketitle
\begin{abstract}
The impressive linguistic abilities of large language models (LLMs) have recommended them as models of human sentence processing, with some conjecturing a positive `quality-power’ relationship, in which language models’ (LMs') fit to psychometric data continues to improve as their ability to predict words in context increases.
This is important because it might suggest that elements of LLM architecture reflect the architecture of the human sentence processing faculty, and that any inadequacies in predicting human reading time and brain imaging data may be attributed to insufficient model complexity, which recedes as larger models become available.
But recent studies have shown this scaling inverts after a point, as LMs become excessively large and accurate, when information-theoretic surprisal is used as a predictor.
Other studies propose the use of entire vectors from differently sized LLMs, still showing positive scaling, casting doubt on the value of surprisal as a predictor, but do not control for dimensionality expansion using untrained LLMs with more than 1.6B parameters.
This study evaluates scaling of LLM vector predictors controlled using untrained LLMs with up to 66B parameters.
Results show that inverse scaling obtains, and moreover the contribution of trained LMs over corresponding untrained LMs drops to zero at around a few billion parameters on most datasets.
\end{abstract}

\section{Introduction and Related Work}
There has been considerable interest in predicting human psychometric data using predictors from LLMs (e.g., \citealt{goodkindbicknell18,schrimpf2021neural,caucheteux2022brains,goldstein2022shared,aw2024instruction,hosseini2024artificial,kauf2024lexical,tuckute2024driving,michaelov2024strong,raugel2025scaling}), with some conjecturing a positive `quality-power' effect of LLM word probabilities on psychometric fit \citep{wilcoxetal23}, favoring larger models trained on more data.
This is important because it might suggest that elements of LLM architecture, such as veridical attention to context and a unique objective of predicting upcoming words, reflect the architecture of the human sentence processing faculty, and that any inadequacies in predicting human psychometric data may be attributed to insufficient model complexity, which recedes as larger models become available.
But the results are mixed.
On the one hand, there have been questions about the linear or superlinear relationship between surprisal and reading time \cite{hooveretal23}, and the predictive power of surprisal on reading times and brain imaging data has been shown to attenuate as successive versions of LLMs predict words with greater accuracy \citep{oh2022comparison,oh2023does,steuer-etal-2023-large,lin2026surprisal}.\footnote{Although studies have shown that surprisal (negative log probability) of words in context is a reliable predictor of reading times \cite{hale01,levy08}.}

On the other hand, recent experiments using vectors from LM layers as predictors (e.g., \citealt{schrimpf2021neural,antonello2023scaling,aw2024instruction,hosseini2024artificial,bonnasse2024fmri}) have shown an improved fit to reading times and brain imaging data as LLMs' ability to predict words improves, supporting the quality-power hypothesis and circumventing some questions about the role of surprisal.
In particular, \citet{schrimpf2021neural} vary the word prediction accuracy of the evaluated models by varying the sizes of the models used in prediction.
However, in moving away from surprisal, these experiments introduce a potential confound in that they not only vary the size (and word prediction accuracy) of the LMs in their studies, they also simultaneously vary the vector size, and thus the number of predictors, which increases the number of degrees of freedom of the regression.
Previous work has shown that large sets of predictors derived from even random high-dimensional models can improve the regression fit \citep{jaeger01,maassetal02,zhang-bowman-2018-language}.
A larger number of predictors gives a regression more degrees of freedom, or more opportunities to find strong predictors.
This is a basis of reservoir computing (RC; \citealt{jaeger01,maassetal02}) in which a single neural layer is trained as a classifier using a larger untrained neural network as input.
RC works by identifying and providing features for latent autoregression in sequence data.
The untrained neural units together serve as a reservoir which captures these autoregressive properties of the input. 

\citet{schrimpf2021neural} consider the larger number of predictors in larger LMs by comparing each trained model with a corresponding untrained version of that model.
However, for decoder-only LMs, which they observe to be the most predictive models of human psychometric data, they do not report results for models larger than 1.6B parameters and do not report on the trends for contributions of trained models over corresponding untrained models as model size increases.
\citet{alkhamissi-etal-2025-language} control for vector size using brain localizer methods \citep{alkhamissi-etal-2025-llm} to select a fixed set of units from the entire network; they find neither positive nor negative scaling of regression fit as the model size increases.
\citet{kuribayashi2025large} further find an improved fit for surprisal at intermediate layer depths, calculated using a logit lens technique \citep{nostalgebraist2020logit}, but do not examine vector predictors.
\citet{feghhi2024large} parcel out the information contained in untrained model representations and find that simple features such as sentence position and sentence length account for the majority of the predictive power of untrained GPT-2 XL, but they do not evaluate on models larger than 1.6B parameters.

This present study extends the set of models of \citet{schrimpf2021neural} to substantially larger decoder-only models with several times more parameters and finds inverse scaling rather than positive scaling.\footnote{This study focuses on evaluating decoder-only LMs, as they are reported to be the most predictive models of human sentence processing \citep{schrimpf2021neural}.}
In addition, this study also extends \citet{alkhamissi-etal-2025-language} by evaluating scaling trends at different layer depths, as model representations from internal layers have been observed to be more predictive than those from the final layers (e.g., \citealt{toneva2019interpreting,antonello2024evidence}), and extends \citet{kuribayashi2025large} by examining vector predictors.
These experiments first replicate the increase in model fit with increasing word probability estimates observed by \citet{schrimpf2021neural} when using vectors from differently sized pre-trained LMs as predictors.
Follow-up experiments which control vector dimensionality expansion using untrained LLM vectors no longer show positive scaling, and indeed show inverse scaling for the trained models when compared against their untrained counterparts, suggesting that the number of predictors is indeed a confound.
Results at different layer depths also show inverse scaling with model size.

When dimensionality expansion is controlled in this way, larger LLMs do not contribute more as they get larger, beyond the effects of degrees of freedom or RC.
Moreover, the contribution of LLM training to model fit over equivalent untrained models at all layer depths drops to zero on most of the datasets tested in this study.
This observed trend away from improving model fit may indicate that there is a substantial misalignment between LLMs and human sentence processing, which worsens when larger models are used, and findings of positive scaling for vectors may instead be due to autoregressive correlations of RC.
At the least, these results argue that untrained versions of LLMs should be used as a standard baseline to evaluate the contribution of LLM training.\footnote{Code for the experiments ran in this study is available at \url{https://github.com/nicalin/llm-vec-dim-control}.}


\section{Experiment 1: Predictive Power of Vectors from Pre-trained LLMs}\label{exp1}
Experiment~1 examines the predictive power of vector elements from 25 pre-trained models from five LM families, GPT-2 \citep{radford2019language}, GPT-Neo \citep{blacketal21, black-etal-2022-gpt, gpt-j}, OPT \citep{zhang2022opt}, Qwen-3 \citep{qwen3technicalreport}, and Gemma-3 \citep{gemma_2025} families, on reading times and brain imaging data.
The vector sizes (number of predictors) of the models evaluated in this experiment vary from 768 (for GPT-2 Small, GPT-Neo 125M, and OPT 125M) to 9,216 (for OPT 66B).

\subsection{Response Data}\label{exp1-resp-data}

This study considers one self-paced reading (SPR), two eye-tracking (ET), and two functional magnetic resonance imaging (fMRI) corpora.

\begin{enumerate}
    \item \citet{futrell2021natural} (\textbf{Natural Stories SPR}): 
    The Natural Stories Corpus includes SPR times from 181 participants who read 10 naturalistic English stories.
    Each participant was asked to answer six comprehension questions after reading each story. 
    \label{natstor-spr}
    
    \item \citet{kennedyetal03} (\textbf{Dundee ET}): 
    The Dundee Corpus includes fixation durations from 10 subjects who read 67 English newspaper articles. 
    \label{dundee-et}
    
    \item \citet{luke2018provo} (\textbf{Provo ET}): 
    The Provo Corpus includes fixation durations from 84 subjects who read 55 short English passages. 
    \label{provo-et}
    
    \item \citet{shain2020fmri} (\textbf{Natural Stories fMRI}):
    The Natural Stories fMRI corpus includes the time series of blood oxygenation level-dependent (BOLD) signals collected with fMRI from the language network \citep{fedorenko2011functional} of 78 subjects, identified using a localizer task.
    The fMRI data, which aggregate BOLD signals across several functional regions of interest (fROIs), were collected at a two-second fixed time interval while the subjects listened to a recording of the Natural Stories Corpus \citep{futrell2021natural}, consisting of naturalistic English stories.
    \label{natstor-fmri}
    
    \item \citet{pereira2018} (\textbf{Pereira fMRI}): 
    The Pereira fMRI corpus includes data from Experiments~II and III in \citet{pereira2018} and the responses are per-sentence BOLD signals. 
    Experiment~II consists of 96 passages (384 sentences in total) read by eight subjects and Experiment~III consists of 72 passages (243 sentences in total) read by six subjects. 
    Five subjects participated in both experiments. 
    For both experiments, subjects were presented with one sentence at a time for four seconds and sentences were separated by a four-second interval. 
    Each passage was presented three times to the subjects.
    To get the average activation of a seeing a sentence, we calculated a BOLD value for each sentence per subject by aggregating activation across regions from the language network.
    \label{pereira}
\end{enumerate}

Among the corpora analyzed in this study, Natural Stories SPR and Pereira fMRI are also analyzed in \citet{schrimpf2021neural}.

\subsection{Data Preprocessing and Partitioning}\label{exp1-data-preprocessing-and-partitioning}

\paragraph{Data Preprocessing.}
The SPR and ET datasets were preprocessed before being divided into different partitions for regression modeling.
The preprocessing procedures we used followed recent work on these datasets \citep{oh2023does}:
For Natural Stories SPR, we excluded data points (1) for sentence-initial and -final words, (2) from participants with fewer than four correct comprehension questions, and (3) outside the [100ms, 3000ms] reading time window.

For both ET datasets, we used first-pass durations as the dependent variable, with data points for unfixated words, words which follow saccades that skip more than four words, sentence-initial/-final words, and document-initial/-final words omitted.
For Dundee ET \citep{kennedyetal03}, which is accompanied by annotations of line and screen positions, data points of words at the boundaries of lines and screens were also filtered out. 

\paragraph{Data Partitioning.}
For all datasets other than Pereira fMRI, we followed the data-partitioning procedure from recent work on these datasets \citep{shain2020fmri,oh-etal-2024-frequency}.
These data were partitioned into fit, exploratory, and held-out sets consisting of roughly 50\%, 25\%, and 25\% of data points after preprocessing.
This resulted in 384,984, 192,826, and 192,449 data points for Natural Stories SPR;  98,115, 48,598, and 48,794 data points for Dundee ET; 52,959, 26,539, and 26,640 for Provo ET; and 100,084, 50,818, and 51,393 data points for Natural Stories fMRI.
For Natural Stories fMRI, for each partition, to ensure sequential time series data in the set and to avoid including data points temporally too close to each other, we grouped the BOLD signals into 30-second chunks and distributed those chunks into the sets.
For Pereira fMRI, due to the small amount of data, we followed \citet{schrimpf2021neural} in the procedures of data-partitioning and regression modeling (Section~\ref{exp1-reg-modeling}), using by-subject five-fold cross-validation when fitting the regression models. 
Following \citet{schrimpf2021neural}, we also treated the data from the two experiments in \citet{pereira2018} separately. 
This resulted in roughly 308 and 76 data points respectively in the per-subject training and test folds for Experiment~II from Pereira fMRI and 195 and 48 data points respectively in those for Experiment~III from Pereira fMRI.%
\footnote{This was the only corpus processed using cross-validation, in order to preserve held-out partitions against overfitting in repeated experiments, since other corpora include a relatively large amount of data points.}

\subsection{Predictors}\label{exp1-predictors}
This study focuses on decoder-only LMs, as they are reported to be the most predictive models of human sentence processing \citep{schrimpf2021neural}.
For this first experiment, which replicates the findings of \citet{schrimpf2021neural}, we collected predictors from 25 models of five autoregressive Transformer-based LM families: GPT-2 \citep{radford2019language}, GPT-Neo \citep{blacketal21, black-etal-2022-gpt, gpt-j}, OPT \citep{zhang2022opt}, Qwen-3 \citep{qwen3technicalreport}, and Gemma-3 \citep{gemma_2025}.
We used all vector elements from the final layer of those LMs as predictors because it is the layer most directly responsible for next-word prediction.
To collect vector elements from the LLMs for the corpora, we first tokenized each story, passage, and article of the corpora with the models' corresponding native tokenizers.
We then input the tokenized texts into each LM to obtain vectors.
In cases where a tokenized story, passage, or article exceeds a single context window of the LMs, to collect vectors for the remaining tokens, we used the latter half of the previous context window as the first half of the context window.

For all SPR and ET datasets, since the response data is a reading time for each word, we obtained the word-level vector by averaging the vectors corresponding to the subword tokens.
For Natural Stories fMRI, since its response data is the time series of BOLD signals, we first collected the vectors and then applied a hemodynamic response function \citep{boynton1996linear} convolution to those vectors so that the convolved vectors align with the response data.
For Pereira fMRI, since its responses are per-sentence BOLD signals, we followed recent work on this dataset by collecting the vector corresponding to the final word of each sentence \citep{hosseini2024artificial}.
If a sentence-final word consists of multiple subword tokens, we average across the vectors of those tokens.
Table~\ref{table:hyper-para} (Appendix~\ref{sec:hyperpara}) details the information about each model's hyperparameters.

\subsection{Regression Modeling}\label{exp1-reg-modeling}
Following \citet{schrimpf2021neural}, to compare the models to the human response data, for each model variant, we fit a linear regression model from the vector elements corresponding to the fit partition of the human response data.
Then, we used the regression model to generate predictions for the held-out partition.\footnote{The exploratory partition was used to explore ways for controlling the dimensionality.}
Also following \citet{schrimpf2021neural}, we examined the predictive power of each model variant by measuring the Pearson's correlation between its predictions and the human responses.
For the SPR and ET datasets, we fit the regression models between the reading times in the fit partition and predictors.
For Natural Stories fMRI, each regression model was fit between the BOLD signals over all fROIs in the fit partition and the predictors.
For Pereira fMRI, as mentioned in Section~\ref{exp1-data-preprocessing-and-partitioning}, we followed \citet{schrimpf2021neural} in the regression modeling procedure by taking the average correlation scores across all five folds of all subjects and across the two experiments.

In addition to vector elements, we also included baseline predictors in the regression.
For Pereira fMRI, we included two baseline predictors -- the position of the sentence in the passage it belongs to and the length of the sentence in words.
The two baselines are reported in \citet{feghhi2024large} to be predictive of the responses in Pereira fMRI.
Since the remaining corpora were not evaluated in \citet{feghhi2024large}, we selected the predictor available in the corpora which is closest to their baselines.
For the remaining corpora, we used the word position in the sentence as the baseline, as they are also a common predictor in psycholinguistic studies.
%
All predictors and responses were centered and scaled before model fitting.

\begin{figure*}[t!]
    \centering
    \includegraphics[width=0.85\textwidth]{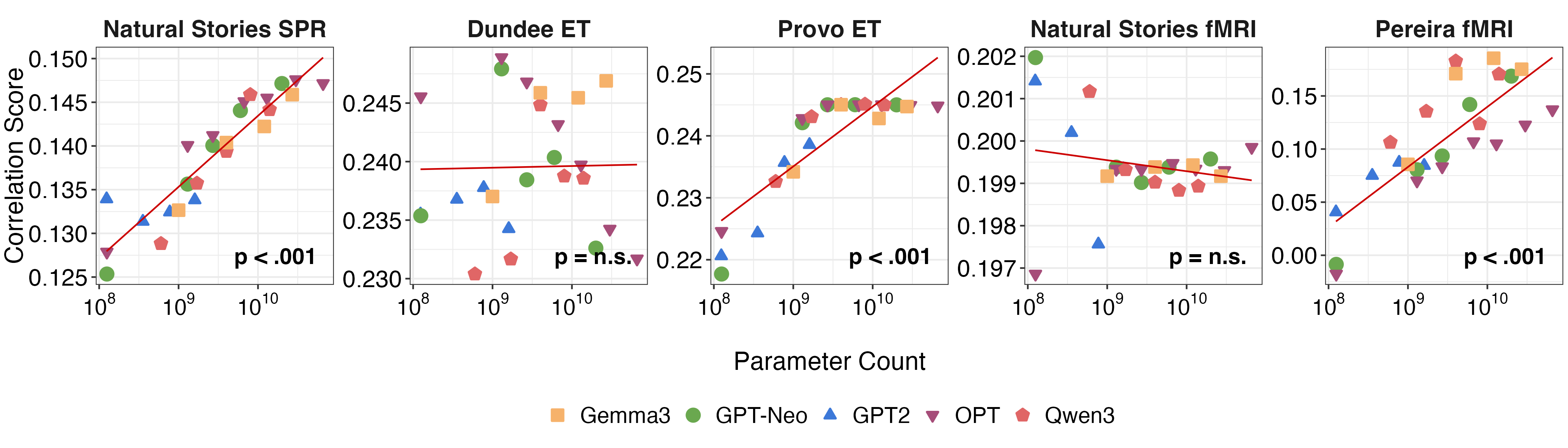}
    \caption{Predictive power of vectors from pre-trained LMs from five LM families on human response data.}
    \label{fig:exp1}
\end{figure*}

\subsection{Results}\label{exp1-results}
Figure~\ref{fig:exp1} shows the results of the relationship between the count of the total number of trainable parameters in each model\footnote{The number of parameters correlates monotonically with word prediction accuracy.} and its ability to predict reading times and BOLD signals.
%
We subtracted the baseline correlation scores from the correlation scores obtained using both baseline predictors and vector elements as predictors.
The baseline correlation scores for Natural Stories SPR, Dundee ET, Provo ET, Natural Stories fMRI, and Pereira fMRI are respectively 0.00418, 0.00926, -0.00978, 0.0389, and 0.149.

The resulting correlation scores across the five LM families and the log-transformed parameter counts were used to obtain a best-fit line, which resulted in lines with slopes significantly greater than 0 for Natural Stories SPR, Provo ET, and Pereira fMRI by a permutation test with 1,000 permutations.
The results of Natural Stories SPR and Pereira fMRI, which are also examined in \citet{schrimpf2021neural}, were replicated and the results of Provo ET are in line with their findings.
Natural Stories fMRI and Dundee ET, however, do not show a significant relationship between predictive power and parameter count in either direction.
Despite this, in the majority of datasets, the results replicate that the model size correlates with quality of fit to reading times and BOLD signals.

\begin{figure*}[t!]
    \centering
    \includegraphics[width=0.85\textwidth]{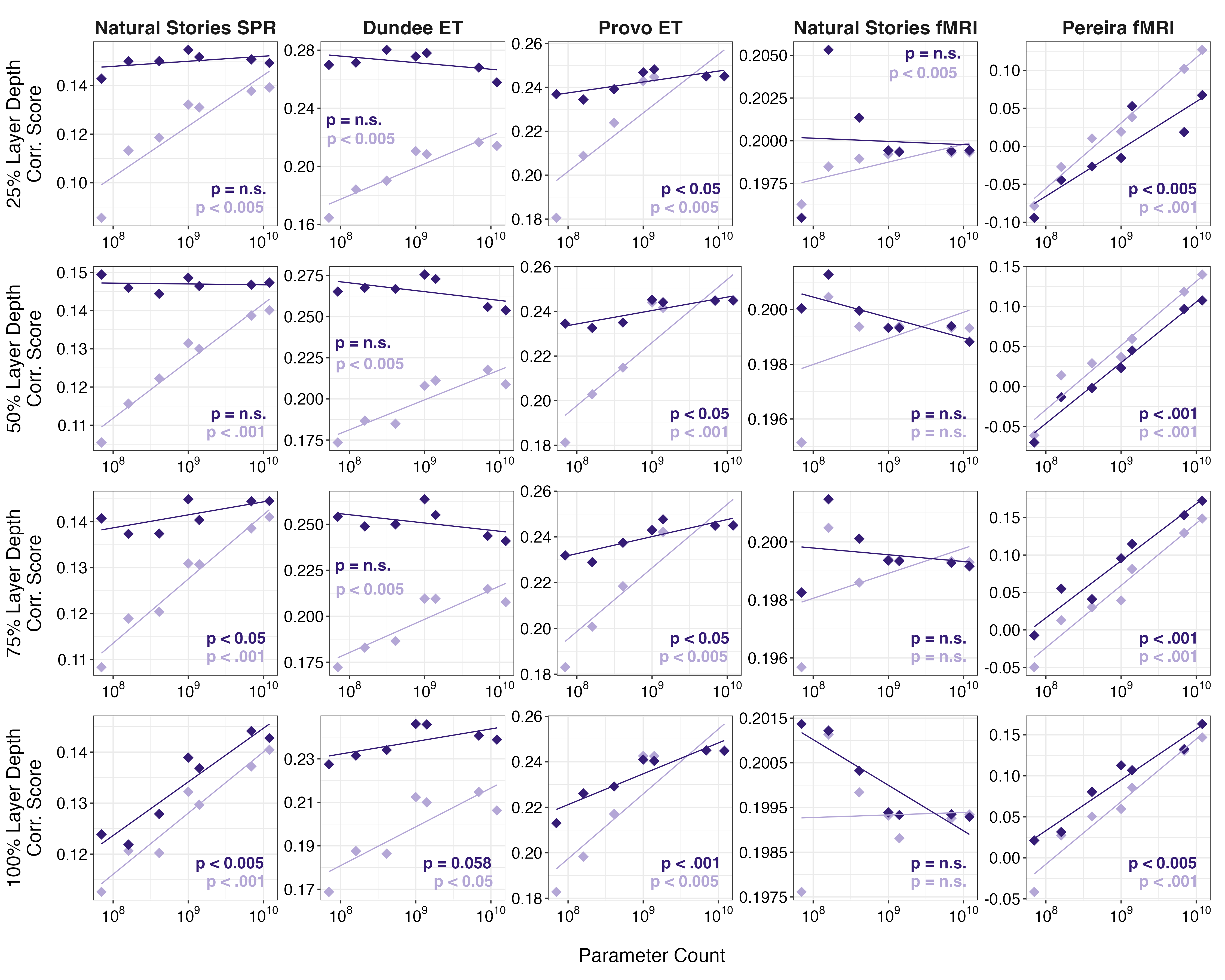}
    \caption{Predictive power of untrained (indicated with light purple) and fully-trained (indicated with dark purple) Pythia model vectors from different layer depths. Results of vectors from low to high percentiles of the layer depth are shown respectively from top to bottom. The significance test of the positive relationship for untrained and fully trained Pythia model vectors is shown respectively in light purple and dark purple text.}
    \label{fig:exp2}
\end{figure*}

\section{Experiment 2: Predictive Power of Vectors from Untrained and Fully Trained LLMs}\label{exp2}

The previous experiment showed a positive quality-power effect when using vector elements directly from fully trained LMs as predictors.
To examine the effect of degrees of freedom or RC of the LMs, we used Pythia models \citep{biderman2023pythia}, as they provide accessibility to various checkpoints during training.
The vector sizes (number of predictors) of the Pythia model variants range from 512 to 5,120.
For each Pythia model variant, we collected vector elements from the model after 0 (untrained) and 143,000 (fully-trained) training steps.
Each training step includes batches of 1,024 samples with 2,048 tokens, resulting in around 300 billion tokens for full training.
We first show the predictive power of the vector elements from untrained and fully trained LMs.
Then, we show the contribution of vector elements from fully trained LMs beyond the effects of degrees of freedom or RC in Section~\ref{contribution}.\footnote{We chose to focus on the Pythia suite for the controlled experiment because it provides access to various checkpoints across training, allowing us to conduct additional analysis of the trajectory of the contribution of LM training throughout the full training process. This additional analysis is shown in Appendix~\ref{sec:training-traj}.}

\subsection{Procedure}

In general, the procedures of this experiment follow those described in Experiment~1, except that we used the Pythia models \citep{biderman2023pythia} and we additionally examined the predictive power of vectors from intermediate layers, as model representations from internal layers have been observed to be more predictive than those from the final layers (e.g., \citealt{toneva2019interpreting,antonello2024evidence}).
Specifically, in order to provide a more general picture of the predictive power of vectors across different layer depths and models with different sizes, we collected vector elements from the layers located at the 25\textsuperscript{th}, 50\textsuperscript{th}, 75\textsuperscript{th}, and 100\textsuperscript{th} (i.e., the final layer) percentiles of each model's depth.
For all corpora, vector elements from untrained and fully trained Pythia variants were collected.
Table \ref{table:hyper-para} in Appendix~\ref{sec:hyperpara} shows the information about each Pythia variant's hyperparameters.
We fit a set of linear regression models from vector elements together with the baseline predictors mentioned in Section~\ref{exp1-reg-modeling} corresponding to the fit partition of each corpus.
The regression models of each corpus were then used to generate predictions for the held-out partition.
The predictive power of all untrained and fully trained Pythia model variants is shown using the Pearson correlation between the predictions and the human response data.

\subsection{Results}
The predictive power of vector elements from untrained and fully trained Pythia variants in the 25\textsuperscript{th}, 50\textsuperscript{th}, 75\textsuperscript{th}, and 100\textsuperscript{th} percentiles of layer depth is shown in Figure~\ref{fig:exp2}.
Similar to the procedure in Experiment~1, for this experiment, we subtracted the baseline correlation scores from those obtained using both baseline predictors and untrained/fully trained LM vectors as predictors.
Figure~\ref{fig:exp2} shows that even when using vector elements from untrained Pythia models, all datasets except for Natural Stories fMRI show a significant positive relationship between predictive power and log-transformed parameter count, regardless of the layer depth.
For Natural Stories fMRI, the positive relationship is significant only when using vectors from the 25\textsuperscript{th} percentile of the layer depth.
These results suggest that the increased vector size of larger models alone already contributes to the better predictive power of larger LMs converging in trained and untrained ones.
As for the fully trained models, the scaling trend is generally consistent with the trend observed for the untrained models but with higher correlation scores compared to the untrained models.

\begin{figure*}[t!]
    \centering
    \includegraphics[width=0.85\textwidth]{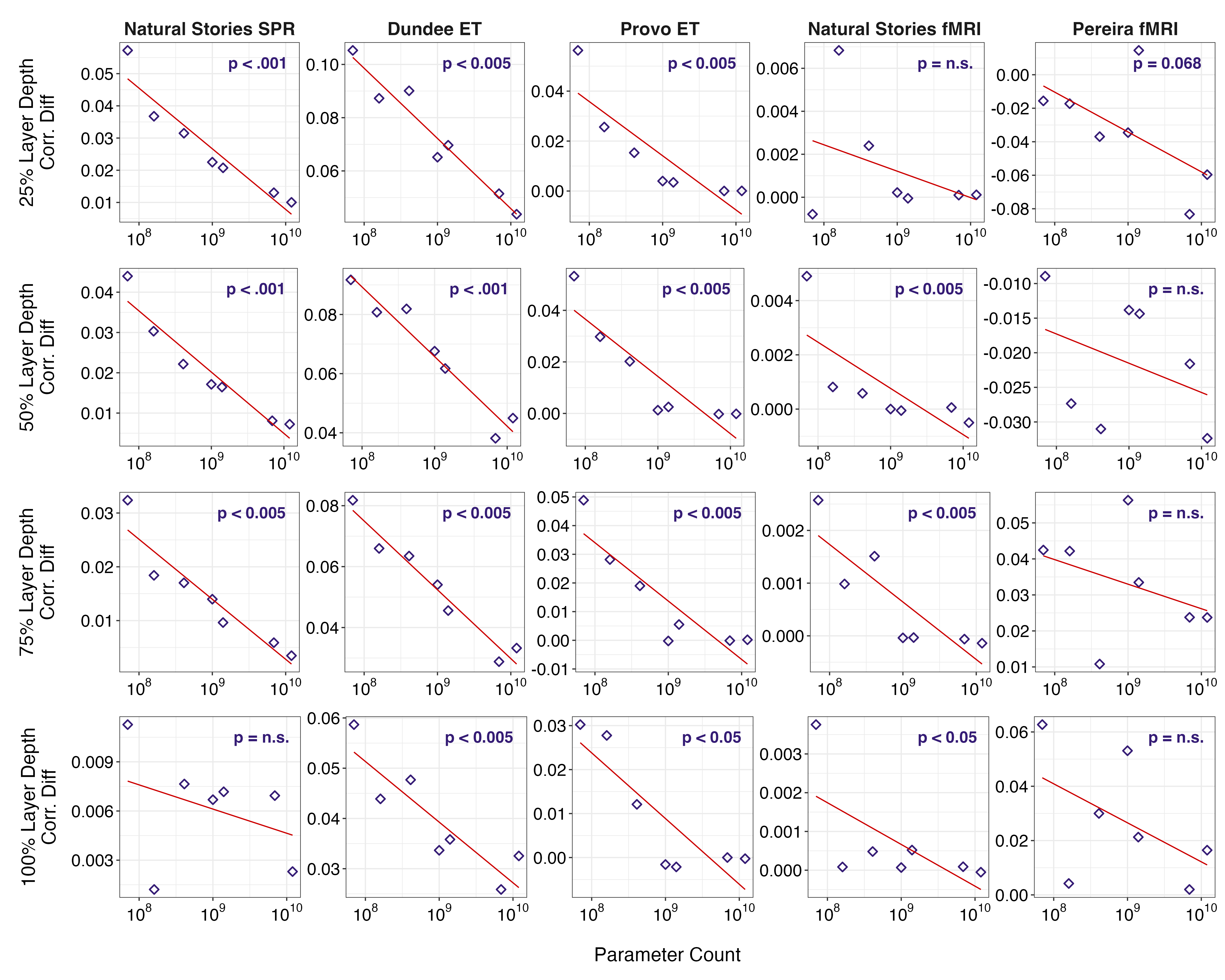}
    \caption{Contribution of additional 143,000 training steps to the predictive power of vector elements from fully trained Pythia models beyond the effect of degrees of freedom or RC. The contribution is calculated by subtracting the predictive power (correlation score) of the untrained LM vector elements from that of the fully trained LM vector elements. Results of vectors from 25\textsuperscript{th}, 50\textsuperscript{th}, 75\textsuperscript{th}, and 100\textsuperscript{th} percentile of the layer depth are shown respectively from top to bottom.}
    \label{fig:exp2-contribution}
\end{figure*}

\section{Contribution of Vectors from Fully Trained LLMs Beyond Effects of RC}\label{contribution}

The previous experiment showed that for the majority of datasets analyzed in this study, the predictive power of vector elements from LMs both without any training and with full training increases as the parameter count (model size) increases, regardless of the layer depth.
To examine how much the full training process of LMs contributes beyond the effect of RC, we subtracted the correlation scores of an untrained Pythia variant from that of its trained counterpart.
Compared to smaller LMs, if the training process of larger LMs contributes more beyond the effect of RC, indicated by the predictive power of vector elements from untrained LMs, larger LMs should have greater correlation score differences than those of smaller LMs.
In this case, greater correlation score differences of larger LMs would suggest that the additional training (i.e., the additional 143,000 training steps) enables vector elements from larger LMs to capture more information about human response data beyond the effect of RC than the information captured by smaller LMs.

Figure~\ref{fig:exp2-contribution} shows that instead of increasing, the correlation score differences of the majority of the corpora decrease significantly as the log-transformed parameter count (model size) increases.
The significant decrease of the correlation score difference is consistently observed across layers from different percentile depths of the models.
14 out of 20 trials show a significant inverse scaling effect.
This overall outcome is strongly significant by a binomial test ($p$~<~.001).
Although some cases show no significant relationship between the correlation score differences and model size (i.e., 100\textsuperscript{th} percentile of the layer depth for Natural Stories SPR, 25\textsuperscript{th} percentile of the layer depth for Natural Stories fMRI, and the layer depths other than the 25\textsuperscript{th} percentile of layer depth for Pereira fMRI), the relationship still shows a numerically negative correlation for those cases.
These results suggest that after (full) training, LMs do not contribute more as they get larger beyond the effect of degrees of freedom or RC.
Moreover, the contribution of LLM training to model fit over equivalent untrained models drops to zero on the Provo ET, Natural Stories fMRI and Pereira fMRI datasets, which constitute a majority of the datasets used in this study.
This may also partially explain the lack of significance of the numerically negative slope observed on Pereira fMRI.\footnote{To ensure that the better predictive power of untrained models is not just specific to the Pythia LM family, we additionally examined the predictive power of untrained models from Experiment 1. Results in Appendix~\ref{sec:exp1-untrained-subtraction} show consistent pattern with what we observed in the Pythia models.}\textsuperscript{,}\footnote{In addition to the untrained and trained Pythia models, we also evaluated the contribution of training across different checkpoints. Results are shown in Appendix~\ref{sec:training-traj}.}

For the main results presented in this study, we followed \citet{schrimpf2021neural} and measured the contribution of vectors from fully trained LMs beyond effects of RC using Pearson's correlation scores from linear regression models.
We also additionally measure the contribution using $R^2$ from the same set of linear regression models and both metrics (i.e., Pearson's correlation score and $R^2$) from ridge regression models.
All additional different types of measurement of the contribution of training beyond effects of RC result in patterns consistent with those in Figure~\ref{fig:exp2-contribution}, showing a significant inverse scaling effect of the contribution of full model training.
Details can be found in Appendices~\ref{contribution-linear-reg-r2} and \ref{contribution-ridge-reg-corr-and-r2}.

Due to noise in the psychometric data, it is possible that the inverse scaling of the contribution of model training is confounded by a ceiling effect.
In this case, the ceiling is the highest possible correlation between psychometric data and predictors derived from other human response data.
A ceiling effect is a potential confound because if the predictive power of larger LMs are approaching the ceiling, there is not much room for the subsequent larger LMs to improve, which can lead to a decrease in the contribution of training in larger models.

To investigate whether the inverse scaling effect of the contribution is confounded by a ceiling effect, we additionally normalized the raw correlation scores from linear and ridge regression models with the ceiling values from \citep{schrimpf2021neural}.
We performed this additional analysis for only Natural Stories SPR and Pereira fMRI, as they are the only two datasets that were also evaluated in \citep{schrimpf2021neural}.
After normalization, the inverse scaling effect still holds for Natural Stories SPR and the normalized correlation scores are far from the upper bound, suggesting that the decrease in contribution of full training in larger models is not due to a ceiling effect.
As for Pereira fMRI, only the normalized predictive power of vectors from trained Pythia 12B at 75\textsuperscript{th} percentile of layer depth reaches the upper bound (i.e., having a correlation score equal to or larger than 1 after normalization); other than this case, the normalized predictive power of vectors from all Pythia models and from different layer depths is below the upper bound.
Since the contribution of training drops to zero even for smaller models and the normalized power of these models is far from the upper bound, a ceiling effect seems unlikely to be a confound for the zero contribution results of Pereira fMRI.
More details are shown in Appendix~\ref{normalization}.

\section{Discussion}
Previous studies have reported inconsistent findings in the relationship between model size and predictive power when using different types of predictors from Transformers.
\citet{oh2022comparison,oh2023does,oh-etal-2024-frequency,lin2026surprisal}, who use surprisal estimates from LLMs as predictors, report an inverse scaling, while \citet{schrimpf2021neural} and others (e.g., \citealt{antonello2023scaling,aw2024instruction,hosseini2024artificial,bonnasse2024fmri}) find a positive quality-power effect when using vector elements from differently sized LMs as predictors.
However, this use of vectors from differently sized LMs introduces a potential confound in that it also simultaneously varies the number of predictors, which increases the number of degrees of freedom of the regression. 
 
Experiments in this paper control for the potential confound of vector size using a richly interconnected neural network (an untrained version of the same Transformer model) and extend the set of LLMs under study, with results instead showing an inverse scaling, and moreover that the contribution of trained LMs over corresponding untrained LMs drops to zero at around a few billion parameters on most datasets.
This suggests that the better predictive power of vector elements from larger LMs is mainly due to the effect of degrees of freedom or RC, which increases with the increasing number of predictors from larger LMs.
The additional training process of the LMs does not contribute more as the model size increases, but instead {\em degrades} the quality of fit, consistent with observations of inverse scaling for word probability \cite{oh2022comparison,oh2023does,lin2026surprisal}.

Experiments in this study use vector elements from differently sized LMs along with baseline predictors motivated by \citet{feghhi2024large} (as described in Section~\ref{exp1-reg-modeling}).
Among all corpora evaluated in this study, Pereira fMRI was also evaluated in \citet{feghhi2024large}.
\citet{feghhi2024large} parcel out the information contained in vectors from untrained GPT-2 XL and find that for Pereira fMRI, sentence position and sentence length account for the majority of the predictive power of untrained GPT-2 XL vectors.
As reported in Section~\ref{exp1-results}, for Pereira fMRI, this study finds that sentence position and sentence length alone yield predictive power comparable to that of the vectors from several of the smallest untrained Pythia models, mostly replicating the results from \citet{feghhi2024large} for sentence position and sentence length on GPT-2-sized models.
However, this study finds that untrained LLMs contribute more to psychometric prediction as model sizes increase beyond the size of GPT-2 XL.

This present study investigated the contribution of training beyond the effects of degrees of freedom or RC by using the untrained LMs as a baseline for their fully trained counterparts.
Recent studies (e.g., \citealt{alkhamissi-etal-2025-language}) investigating the model-to-brain alignment across the training trajectory of LMs have found that LMs best align to brain imaging data in early training; the alignment then saturates or declines in subsequent training.
To examine whether our observation using untrained and fully trained LMs holds for the partially trained LMs, we also examined dynamic of the contribution of training across the training trajectory.
We focused on vectors from the 25\textsuperscript{th} percentile of the layer depth, as it is shown in Figure~\ref{fig:exp2} to be in general the most predictive layer depth compared to other depths.
Consistent with the finding from \citet{alkhamissi-etal-2025-language}, in general, the contribution of training to the predictive power was also observed to saturate early on in training for all LMs, regardless of their sizes.
In addition, results of this follow-up analysis also show consistent results with those of Experiment~2 in that smaller LMs benefit more from training.
More details of this analysis can be found in Appendix~\ref{sec:training-traj}.

Finally, it may be tempting to interpret the positive effect for larger untrained LMs as showing some inherent advantage of a Transformer architecture, as \citet{schrimpf2021neural} do.
However, it is important to point out that when untrained, these models are not calculating predictivity-based similarity or attention scores over their contexts, and have no objective of next word prediction, which are the properties that make Transformers successful models of language; we suggest that they are simply heavily interconnected neural networks, essentially just suitably autoregressive to perform RC.\footnote{Indeed, reservoir models were proposed as a model of cortex in part because they do not require backpropagation, which has not been observed in biological neurons \citep{crick1989recent}.}

\section{Conclusion}
This work evaluates the number of predictors as a potential confound to the quality-power effect when using vector elements directly from differently sized LMs as predictors.
Results show that, when dimensionality expansion is controlled using untrained LLMs, larger LMs no longer contribute more beyond the effect of simple RC.
This observation suggests that there may be a substantial misalignment between LLMs and human sentence processing, which worsens when larger models are used.
Moreover, these results show the contribution of trained LLMs over untrained LLMs drops to zero on most evaluated datasets after model size exceeds approximately a few billion parameters, suggesting that LLM training may have no substantial correlation to human sentence processing in the limit, beyond the effect of RC.
Indeed, these results appear to recommend using very large untrained reservoir models (possibly other than Transformers) instead of expensively trained LLMs as models of expectation in sentence processing.
At the least, these results argue that untrained versions of LLMs should be used as a standard baseline to evaluate the contribution of LLM training.

\section*{Limitations}
Experiments described in this study attempt to identify a confound of using vectors from differently sized LLMs as predictors to predict human sentence processing data. 
The datasets evaluated are collected from English speakers.
\citet{schrimpf2021neural} claim that there is a positive scaling on English behavioral and neural data.
A finding of inverse scaling on English should therefore be sufficient to contradict the claim.
However, since this study focuses on English data, these findings may or may not be replicated in other languages.

\section*{Ethics Statement}
The datasets used in this study are from published work \citep{futrell2021natural,kennedyetal03,luke2018provo,shain2020fmri,pereira2018}.
These datasets were used to examine LLMs' ability to predict human reading times and brain imaging data.
The focus of this work is to identify a confound in the inconsistent findings in the quality-power hypothesis in previous studies.
The potential risks and harmful impacts posed by this study on the society appear to be minimal.

\section*{Acknowledgments}

We thank the anonymous reviewers for their helpful feedback. This work was partially supported by the National Science Foundation grant \#2313140. All views expressed are those of the authors and do not necessarily reflect the views of the National Science Foundation.

\bibliography{custom}

\appendix
\clearpage


\section{Hyperparameters of LMs}\label{sec:hyperpara}
The hyperparameters of the LMs, grouped in terms of their LM families are shown in Table~\ref{table:hyper-para}.
All models evaluated in this study are pre-trained and without any fine-tuning.
The embedding sizes of the models evaluated in this study are consistent across layers.
The LMs evaluated in this study are autoregressive LMs which can be loaded using the HuggingFace Transformers library.\footnote{\url{https://huggingface.co/docs/transformers/index}}

\begin{table}[h]
\setlength{\tabcolsep}{4pt}
\centering
\begin{tabular}{lrrrr}
\thickhline
\textbf{Model Variant} & \textbf{\#L} & \textbf{\#H} & \textbf{$d_{model}$} & \textbf{Parameters} \\ \thickhline
GPT-2 Small            & 12           & 12           & 768        & $\sim$124M          \\
GPT-2 Medium           & 24           & 16           & 1024       & $\sim$355M          \\
GPT-2 Large            & 36           & 20           & 1280       & $\sim$774M          \\
GPT-2 XL               & 48           & 25           & 1600       & $\sim$1.6B          \\ \hline
GPT-Neo 125M           & 12           & 12           & 768        & $\sim$125M          \\
GPT-Neo 1.3B           & 24           & 16           & 2048       & $\sim$1.3B          \\
GPT-Neo 2.7B           & 32           & 20           & 2560       & $\sim$2.7B          \\
GPT-J 6B               & 28           & 16           & 4096       & $\sim$6B            \\
GPT-NeoX 20B           & 44           & 64           & 6144       & $\sim$20B           \\ \hline
OPT 125M               & 12           & 12           & 768        & $\sim$125M          \\
OPT 1.3B               & 24           & 32           & 2048       & $\sim$1.3B          \\
OPT 2.7B               & 32           & 32           & 2560       & $\sim$2.7B          \\
OPT 6.7B               & 32           & 32           & 4096       & $\sim$6.7B          \\
OPT 13B                & 40           & 40           & 5120       & $\sim$13B           \\
OPT 30B                & 48           & 56           & 7168       & $\sim$30B           \\
OPT 66B                & 64           & 72           & 9216       & $\sim$66B           \\ \hline
Qwen-3 0.6B            & 28           & 16           & 1024       & $\sim$600M          \\
Qwen-3 1.7B            & 28           & 16           & 2048       & $\sim$1.7B          \\
Qwen-3 4B              & 36           & 32           & 2560       & $\sim$4B            \\
Qwen-3 8B              & 36           & 32           & 4096       & $\sim$8B            \\
Qwen-3 14B             & 40           & 40           & 5120       & $\sim$14B           \\ \hline
Gemma-3 1B             & 26           & 4            & 1152       & $\sim$1B            \\
Gemma-3 4B             & 34           & 8            & 2560       & $\sim$4B            \\
Gemma-3 12B            & 48           & 16           & 3840       & $\sim$12B           \\
Gemma-3 27B            & 62           & 32           & 5376       & $\sim$27B           \\ \hline
Pythia 70M             & 6            & 8            & 512        & $\sim$70M           \\
Pythia 160M            & 12           & 12           & 768        & $\sim$160M          \\
Pythia 410M            & 24           & 16           & 1024       & $\sim$410M          \\
Pythia 1B              & 16           & 8            & 2048       & $\sim$1B            \\
Pythia 1.4B            & 24           & 16           & 2048       & $\sim$1.4B          \\
Pythia 6.9B            & 32           & 32           & 4096       & $\sim$6.9B          \\
Pythia 12B             & 36           & 40           & 5120       & $\sim$12B           \\ \thickhline
\end{tabular}
\caption{Hyperparameters of the models examined in this study. \#L refers to the number of layers of that model; \#H refers to the number of attention heads each layer has; $d_{model}$ refers to the embedding size of the model.}
\label{table:hyper-para}
\end{table}


\section{Predictive Power of Untrained Models in Experiment~1 and Contribution of Training beyond Effect of RC}\label{sec:exp1-untrained-subtraction}
\begin{figure*}[t!]
    \centering
    \includegraphics[width=0.85\textwidth]{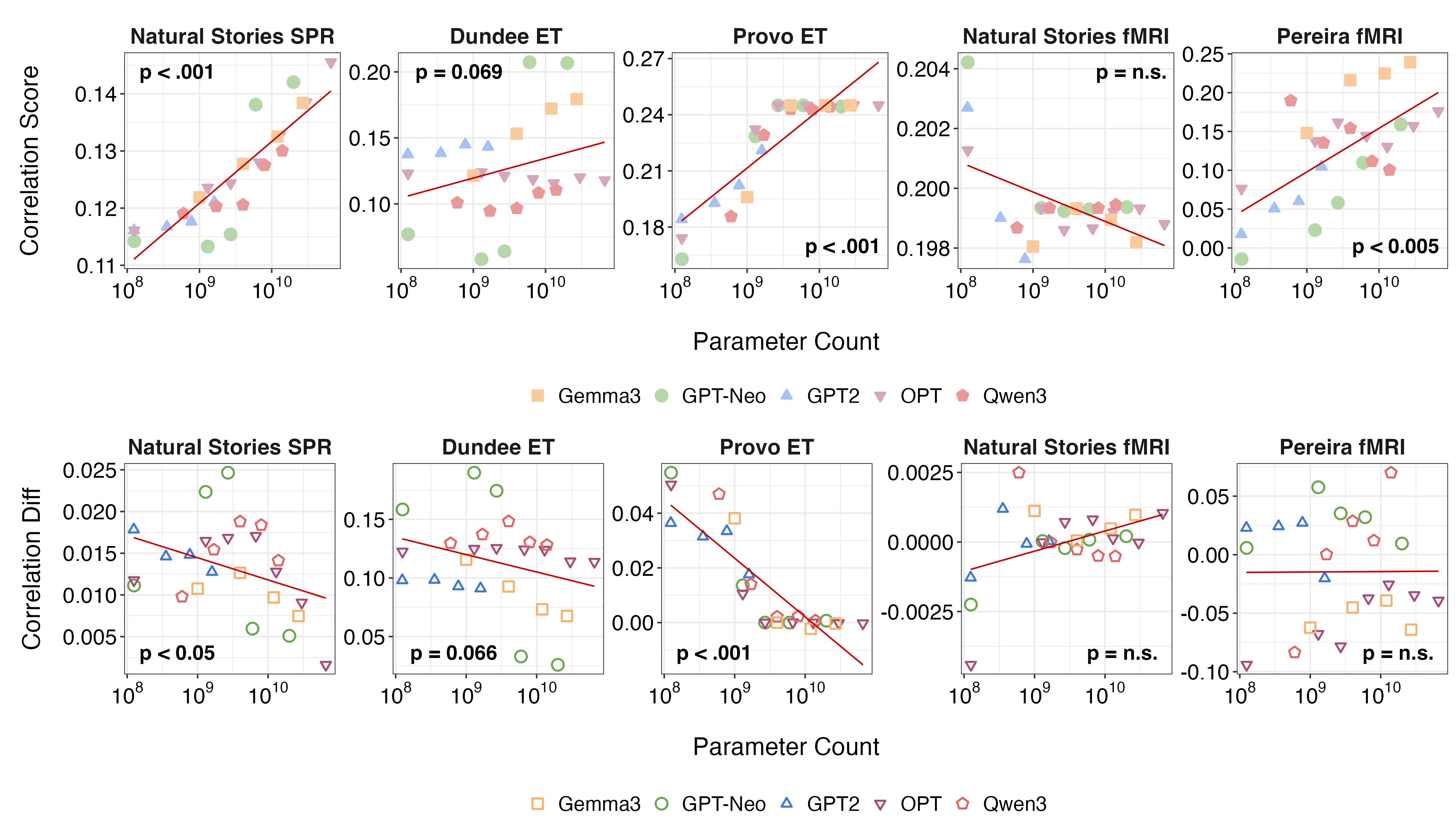}
    \caption{Predictive power of untrained models from Experiment~1 (top) and the contribution of training to the predictive power of vector elements from fully trained models beyond the effect of degrees of freedom or RC (bottom).}
    \label{fig:exp1-untrained-subtraction}
\end{figure*}

To ensure that the better predictive power of untrained models is not just specific to the Pythia LM family, we additionally examined the predictive power of untrained version of the models from Experiment~1. We used the same random seed for all model initialization and used vectors from the final layer, following the procedure for Experiment~1. Figure~\ref{fig:exp1-untrained-subtraction} (top) shows results consistent with those of Pythia models. Natural Stories SPR, Provo ET, and Pereira fMRI show a significant positive relationship between predictive power and model size. The positive relationship for Dundee ET is weakly significant.

We then subtracted the correlation scores of the untrained models from those of the pre-trained ones reported in Experiment~1. Results in Figure~\ref{fig:exp1-untrained-subtraction} (bottom) replicate our finding of the contribution of trained over untrained models reported in Section~\ref{contribution}, which trends to zero on Provo ET, Natural Stories fMRI and Pereira fMRI. The significant inverse relationship between model size and the contribution of trained LMs over untrained ones for Natural Stories SPR and Provo ET are also replicated. Dundee ET in this analysis shows a weakly significant inverse relationship, which is still consistent with the results shown in Section~\ref{contribution}.

\newpage


\section{The Contribution of Training Across Training Trajectory}\label{sec:training-traj}
\begin{figure*}[t!]
    \centering
    \includegraphics[width=0.75\textwidth]{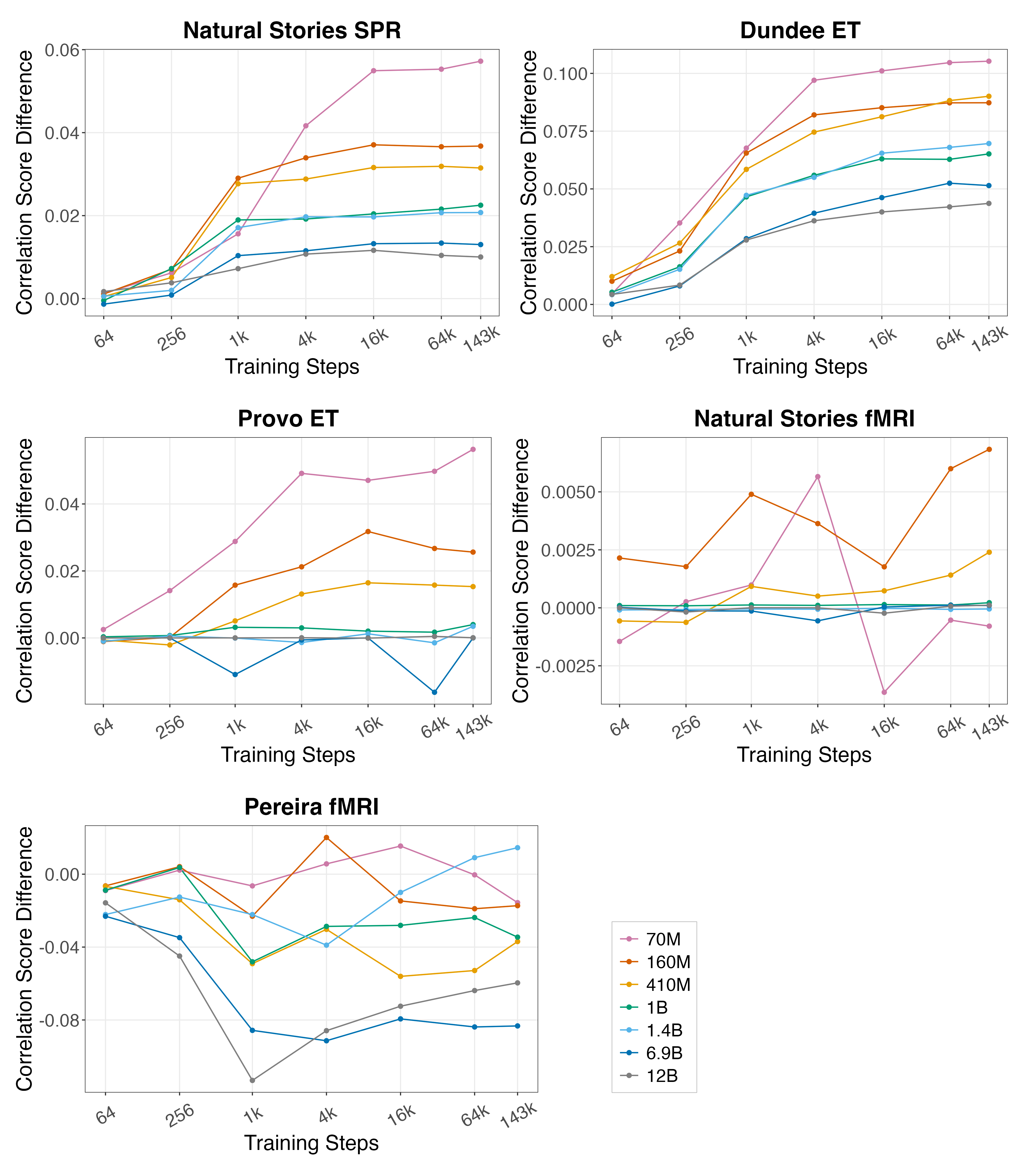}
    \caption{Contribution of training across the training trajectory to the predictive power of vector elements from the 25\textsuperscript{th} percentile of layer depth. The contribution was examined by subtracting the correlation score of step 0 (untrained) from that of the intermediate training steps up until the models were fully trained (step 143,000).}
    \label{fig:trajectory}
\end{figure*}

\citet{alkhamissi-etal-2025-language} observed that the model-to-brain alignment peaks early on in training and then subsequently saturates.
As a follow-up analysis to our Experiment~2, we explored the contribution of training across the training process.
That is, we examined how much more predictive the vectors become at intermediate checkpoints throughout the course of training, compared to vectors from untrained LMs.
Similar to our procedure described in Section~\ref{contribution}, we measured the contribution of training at each checkpoint by subtracting the correlation scores of the untrained LMs (step 0) from the correlation scores of LMs at that checkpoint.
If the training process does continue to contribute more to the predictive power of the LMs, we should observe the correlation difference to continue growing.
In this analysis, we focused on vectors from the 25\textsuperscript{th} percentile of the layer depth, as it is shown in Figure~\ref{fig:exp2} to be in general the most predictive layer depth compared to other depths.
The checkpoints we include are those in the early and late training as well as those around 2B to 8B tokens (i.e., step 1,000 to step 4,000), which were checkpoints reported to start showing saturation \citep{alkhamissi-etal-2025-language}.
Our results shown in Figure~\ref{fig:trajectory} are generally consistent with the result from \citet{alkhamissi-etal-2025-language} in that we also observed a saturation of contribution of training at around step 1,000 to step 4,000 for the majority of datasets.
Moreover, these results are in general consistent with the results reported in Section~\ref{contribution} in that smaller LMs benefit more from training.

\section{Contribution of Vectors from Fully Trained LLMs Beyond Effects of RC Measured Using $R^{2}$ from Linear Regression Models}\label{contribution-linear-reg-r2}

\begin{figure*}[t!]
    \centering
    \includegraphics[width=0.76\textwidth]{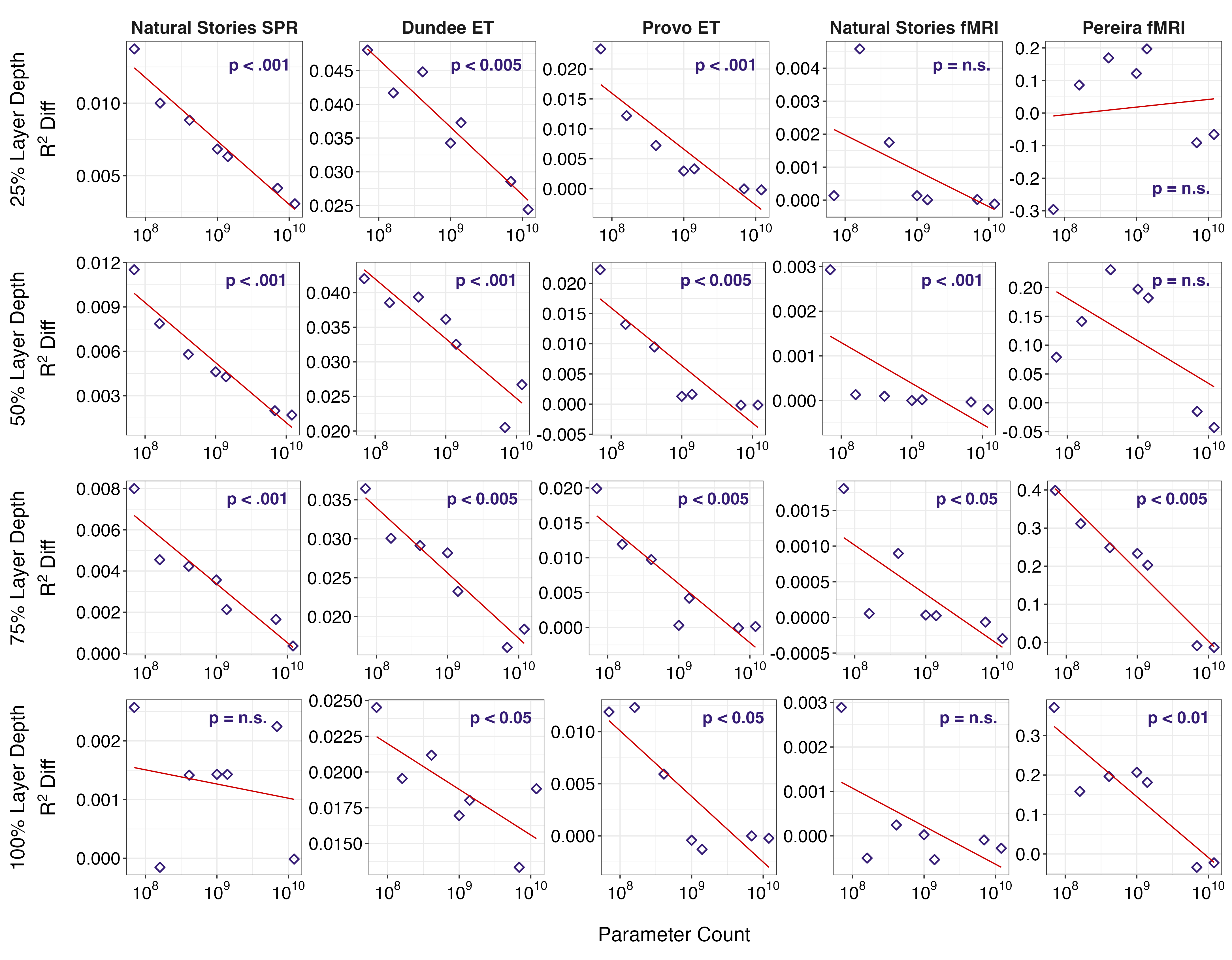}
    \caption{Contribution (measured using $R^2$ from linear regression models) of full model training to the predictive power of vector elements from fully trained Pythia models beyond the effect of degrees of freedom or RC. The contribution is calculated by subtracting the $R^2$ values from linear regression models fit using untrained LM vector elements from $R^2$ values from regression models fit using fully trained LM vector elements.}
    \label{fig:lienar-reg-r2-subtraction}
\end{figure*}

In addition to Pearson's correlation scores, we also measure the predictive power of LM using $R^2$ from linear regression models.
Figure~\ref{fig:lienar-reg-r2-subtraction} shows that when using $R^2$ from linear regression models, the inverse scaling of the difference between fully trained and untrained Pythia models still holds.
All but one trial show numerically inverse scaling and 15 out of 20 trials show significant inverse scaling.
This overall outcome is strongly significant by a binomial test ($p$~<~.001).

\section{Contribution of Vectors from Fully Trained LLMs Beyond Effects of RC from Ridge Regression Models}\label{contribution-ridge-reg-corr-and-r2}

\begin{figure*}[h!]
    \centering
    \includegraphics[width=0.76\textwidth]{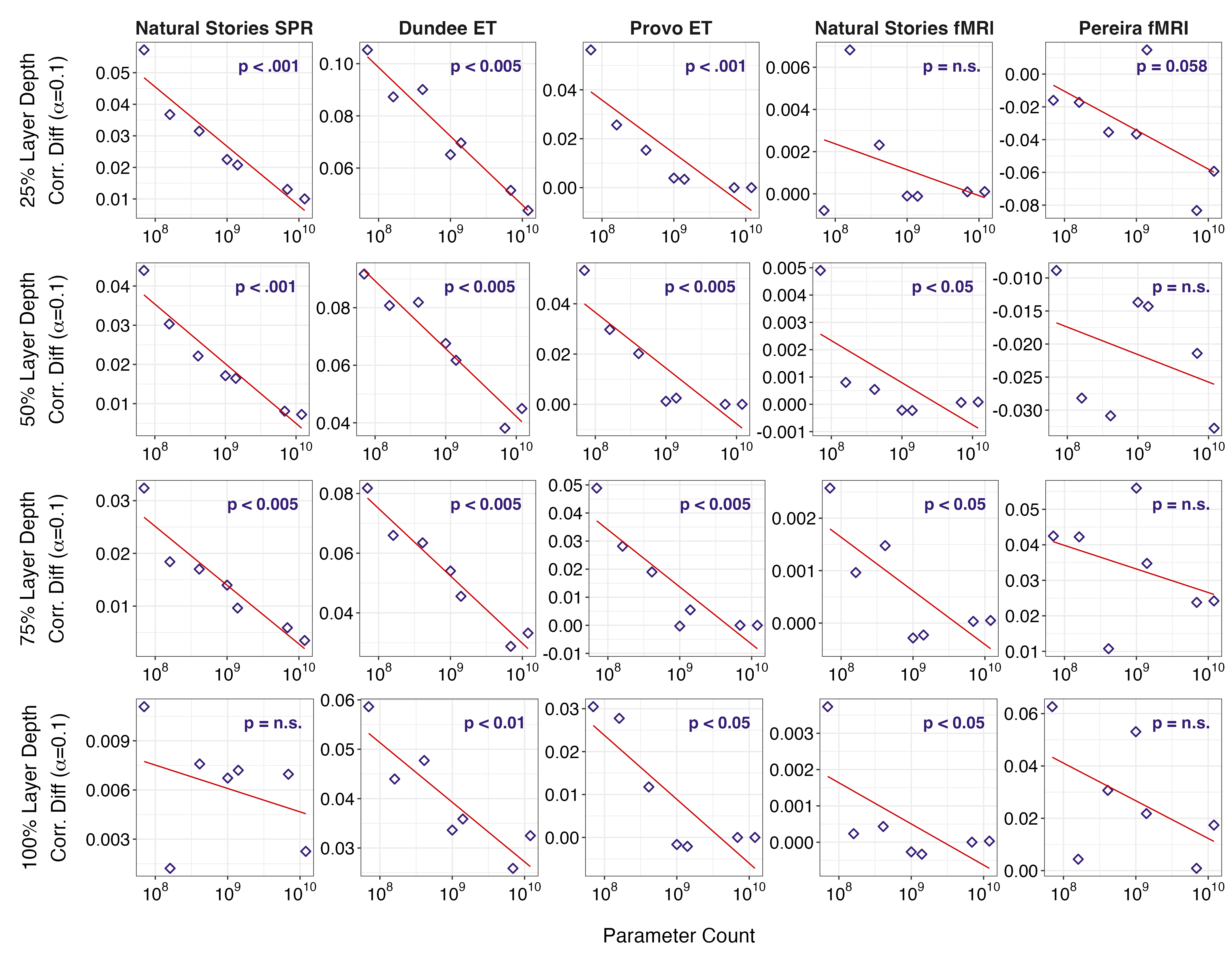}
    \caption{Contribution full training to the predictive power of vector elements from fully trained Pythia models beyond the effect of degrees of freedom or RC. The results reported here are difference in Pearson's correlation scores from the set of ridge regression ($\alpha$~=~0.1) models fit using vectors from untrained and fully trained Pythia models.}
    \label{fig:ridge-alpha0.1-unnormalized-corr-subtraction}
\end{figure*}

\begin{figure*}[h!]
    \centering
    \includegraphics[width=0.76\textwidth]{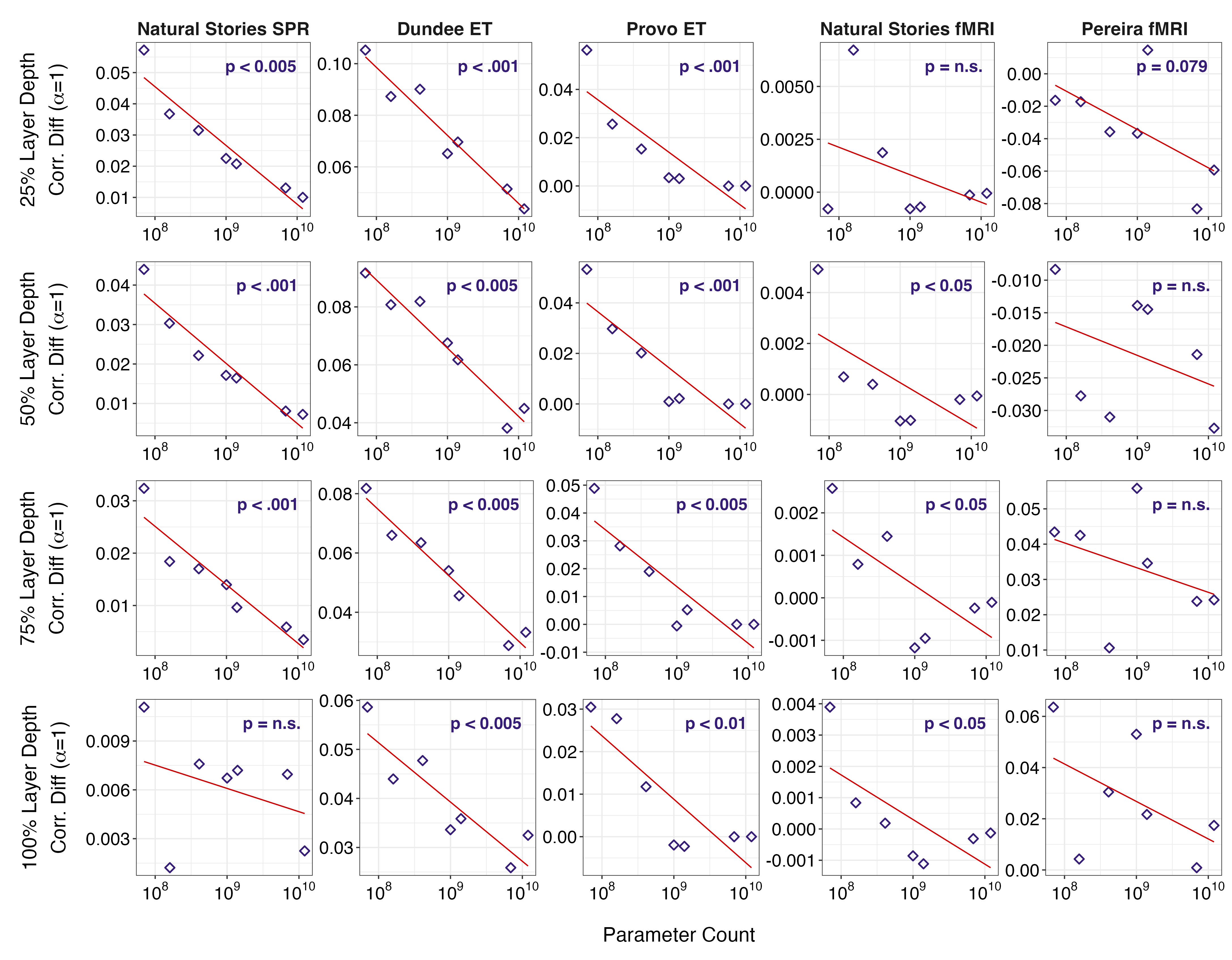}
    \caption{Contribution full training to the predictive power of vector elements from fully trained Pythia models beyond the effect of degrees of freedom or RC. The results reported here are difference in Pearson's correlation scores from the set of ridge regression ($\alpha$ = 1) models fit using vectors from untrained and fully trained Pythia models.}
    \label{fig:ridge-alpha1-unnormalized-corr-subtraction}
\end{figure*}

In addition to linear regression, we also fit a set of ridge regression models which have explicit regularization.
We ran ridge regression using two $\alpha$ values, 0.1 and 1.
Similar to the procedure described in Section~\ref{contribution}, we subtracted the Pearson's correlation scores of untrained Pythia models from those of fully trained Pythia models.
All trials in Figures~\ref{fig:ridge-alpha0.1-unnormalized-corr-subtraction} and \ref{fig:ridge-alpha1-unnormalized-corr-subtraction} show numerically inverse scaling.
14 out of 20 were statistically significant, regardless of the $\alpha$ values we used.
This overall outcome is strongly significant by a binomial test ($p$~<~.001).

\begin{figure*}[h!]
    \centering
    \includegraphics[width=0.76\textwidth]{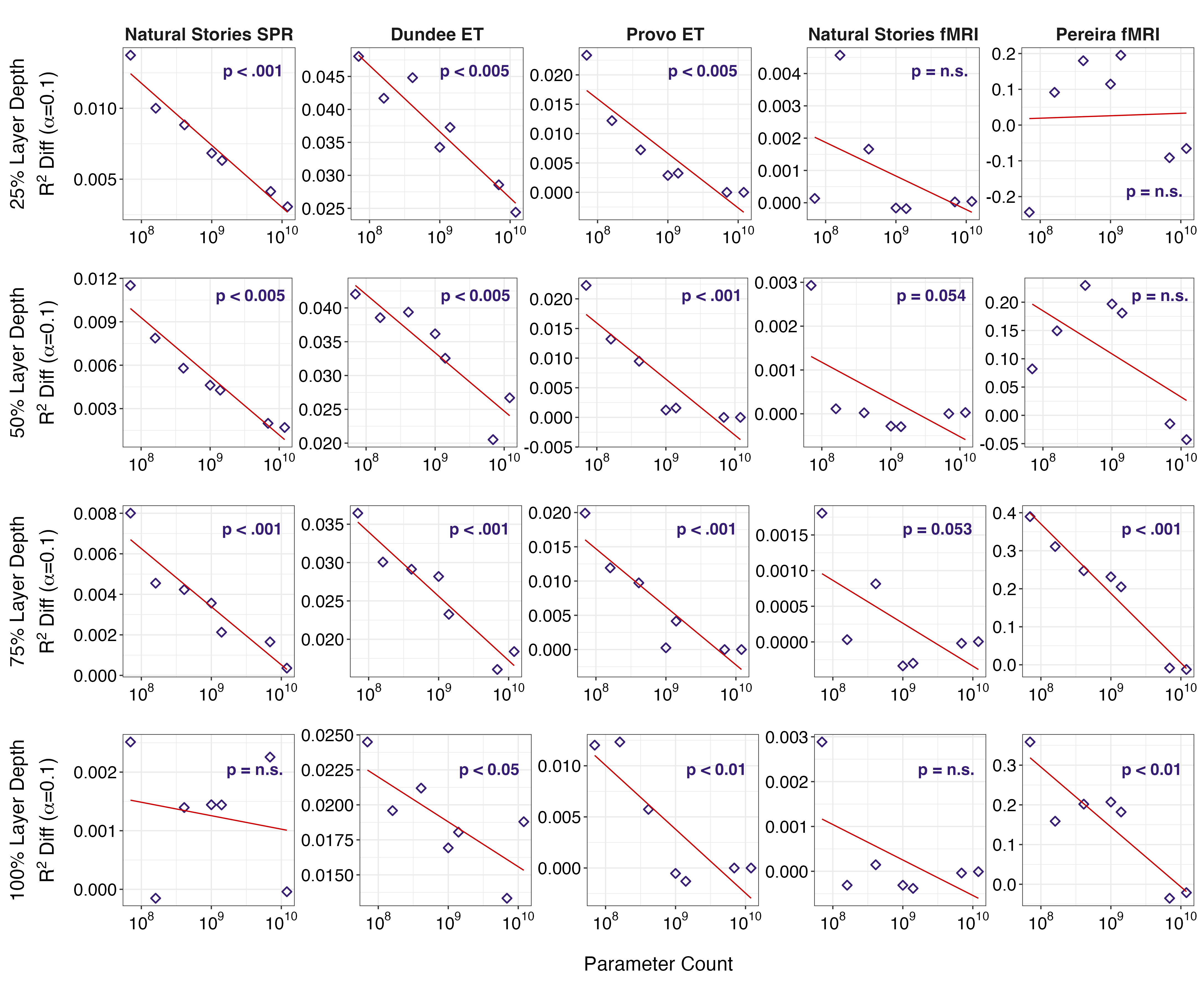}
    \caption{Contribution (measured using $R^2$ from ridge regression models) of full model training to the predictive power of vector elements from fully trained Pythia models beyond the effect of degrees of freedom or RC. The contribution is calculated by subtracting the $R^2$ values from ridge regression models ($\alpha$~=~0.1) fit using untrained LM vector elements from $R^2$ values from regression models fit using fully trained LM vector elements.}
    \label{fig:ridge-alpha0.1-r2-subtraction}
\end{figure*}

\begin{figure*}[h!]
    \centering
    \includegraphics[width=0.76\textwidth]{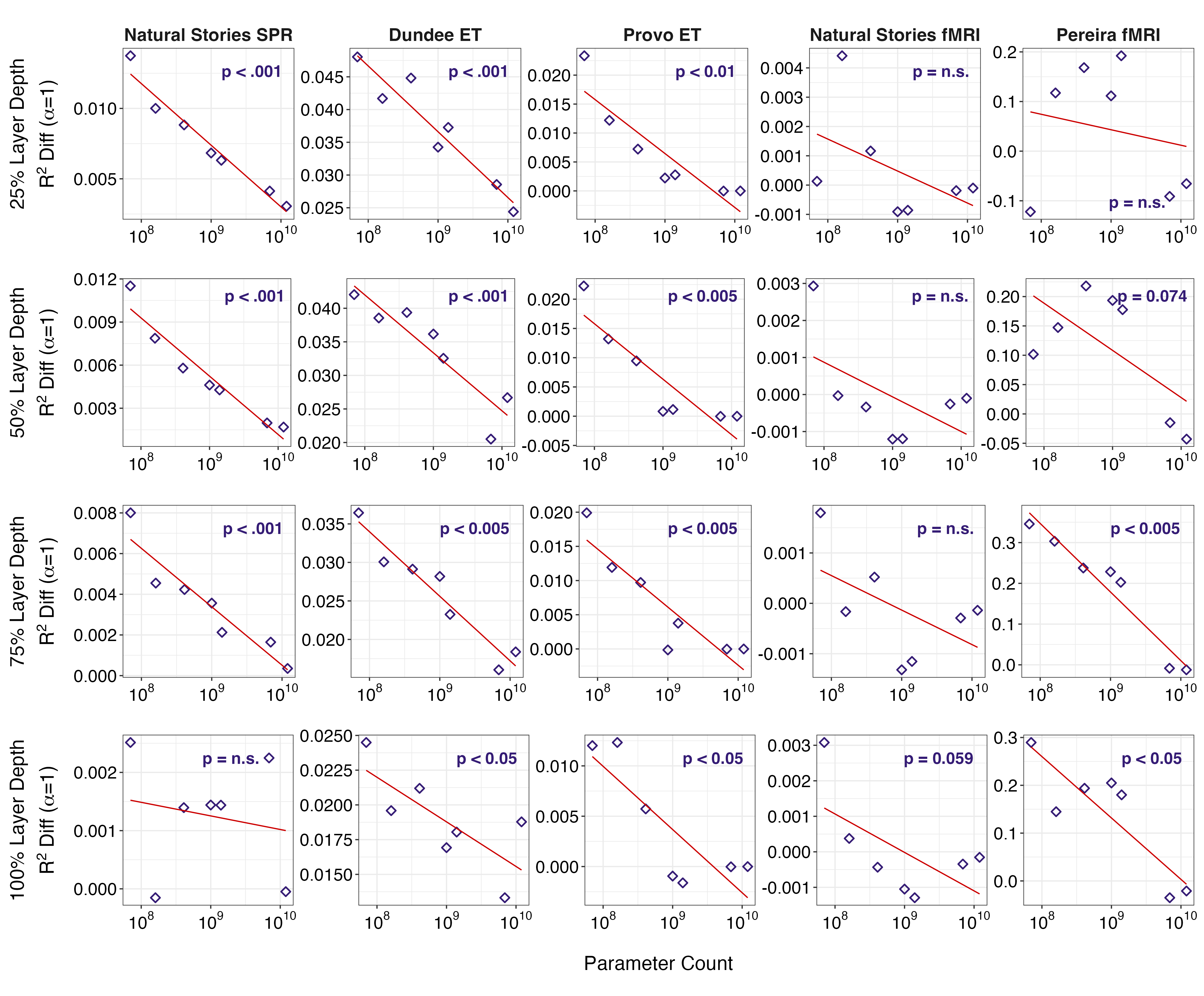}
    \caption{Contribution (measured using $R^2$ from ridge regression models) of full model training to the predictive power of vector elements from fully trained Pythia models beyond the effect of degrees of freedom or RC. The contribution is calculated by subtracting the $R^2$ values from ridge regression models ($\alpha$~=~1) fit using untrained LM vector elements from $R^2$ values from regression models fit using fully trained LM vector elements.}
    \label{fig:ridge-alpha1-r2-subtraction}
\end{figure*}

Similar to the analysis presented in Appendix~\ref{contribution-linear-reg-r2}, we also report the $R^2$ difference between untrained and fully trained LMs from a set of ridge regression models with two $\alpha$ values, 0.1 and 1 (Figures~\ref{fig:ridge-alpha0.1-r2-subtraction} and \ref{fig:ridge-alpha1-r2-subtraction}).
When setting $\alpha$ to 0.1, all but one trial show numerically inverse scaling; when setting $\alpha$ to 1, all trials show numerically inverse scaling.
13 out of 20 were statistically significant, regardless of the $\alpha$ values we used.
This overall outcome is strongly significant by a binomial test ($p$~<~.001).

\section{Predictive of Vectors from Untrained and Fully Trained LLMs after Normalization}\label{normalization}

To investigate whether the inverse scaling effect of the contribution is confounded by a ceiling effect, we additionally normalized the raw correlation scores from linear and ridge regression models with the ceiling values from \citet{schrimpf2021neural}.
In this case, the ceiling is the highest possible correlation between psychometric data and predictors derived from other human response data.
A ceiling effect is a potential confound because if the predictive power of larger LMs are approaching the ceiling, there is not much room for the subsequent larger LMs to improve, which can lead to a decrease in the contribution of training in larger models.
Since Natural Stories SPR and Pereira fMRI are the only two datasets which are also evaluated \citet{schrimpf2021neural}, we only ran this additional analysis on those two datasets.
The ceiling values for Natural Stories SPR and Pereira fMRI are respectively 0.76 and 0.32.
The raw Pearson's correlation scores were normalized by being divided by the ceiling values.
Note that since the purpose of this additional analysis is to show whether the full regression model prediction reaches the ceiling, we did not subtract the baseline correlation scores from correlation scores from regression models fit using both baseline predictors and the vectors.
The raw correlation scores used in this analysis are therefore higher than those reported in Section~\ref{exp2}.

\begin{figure}[t!]
    \centering
    \includegraphics[width=0.85\columnwidth]{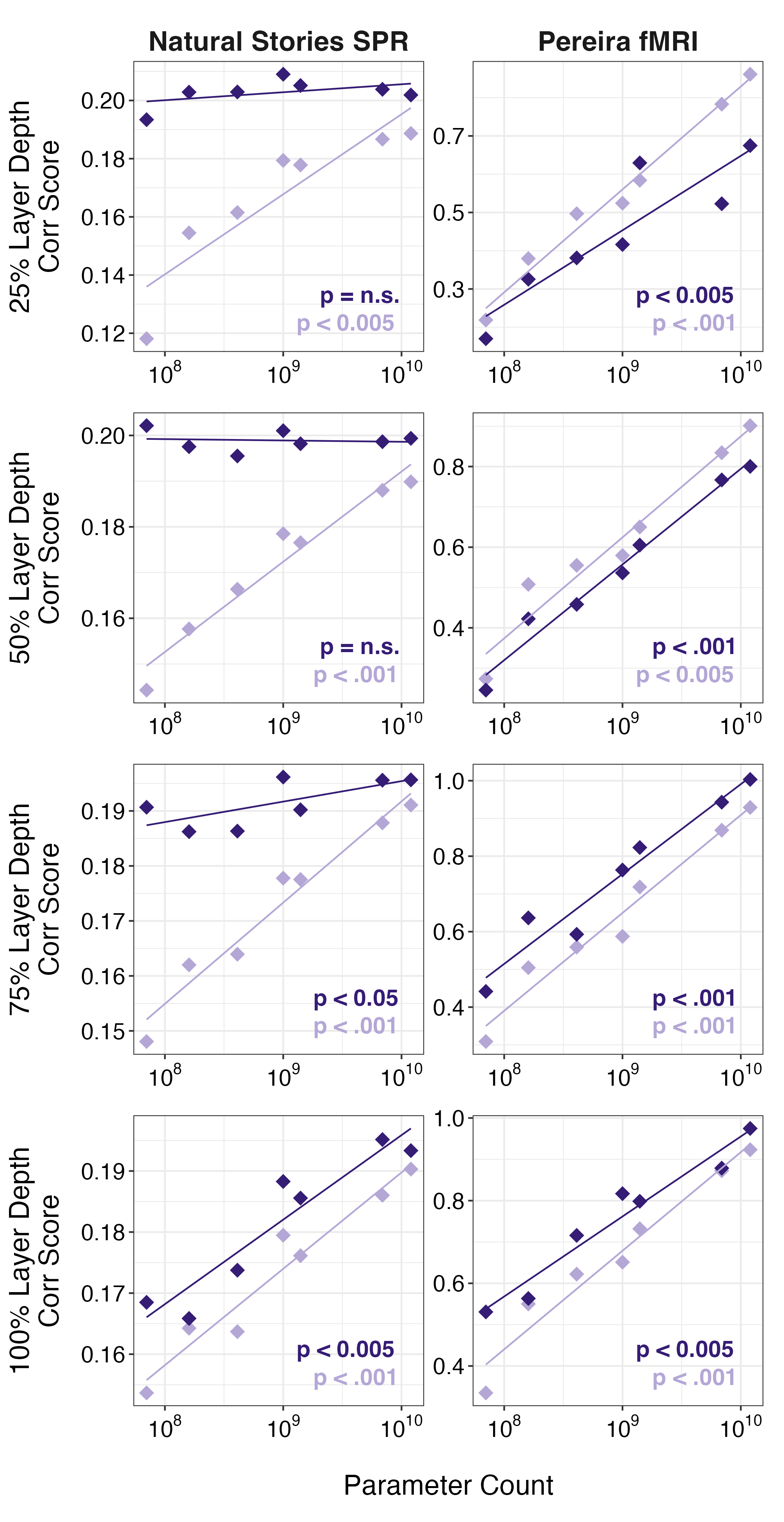}
    \caption{Predictive power (Pearson's correlation scores) of untrained (indicated with light purple) and fully-trained (indicated with dark purple) Pythia model vectors from different layer depths. The correlation scores reported in this figure are from linear regression models and are normalized with the ceiling values provided in \citet{schrimpf2021neural}.}
    \label{fig:linear-reg-normalized-corr-without-baseline-sub}
\end{figure}

\begin{figure}[t!]
    \centering
    \includegraphics[width=0.9\columnwidth]{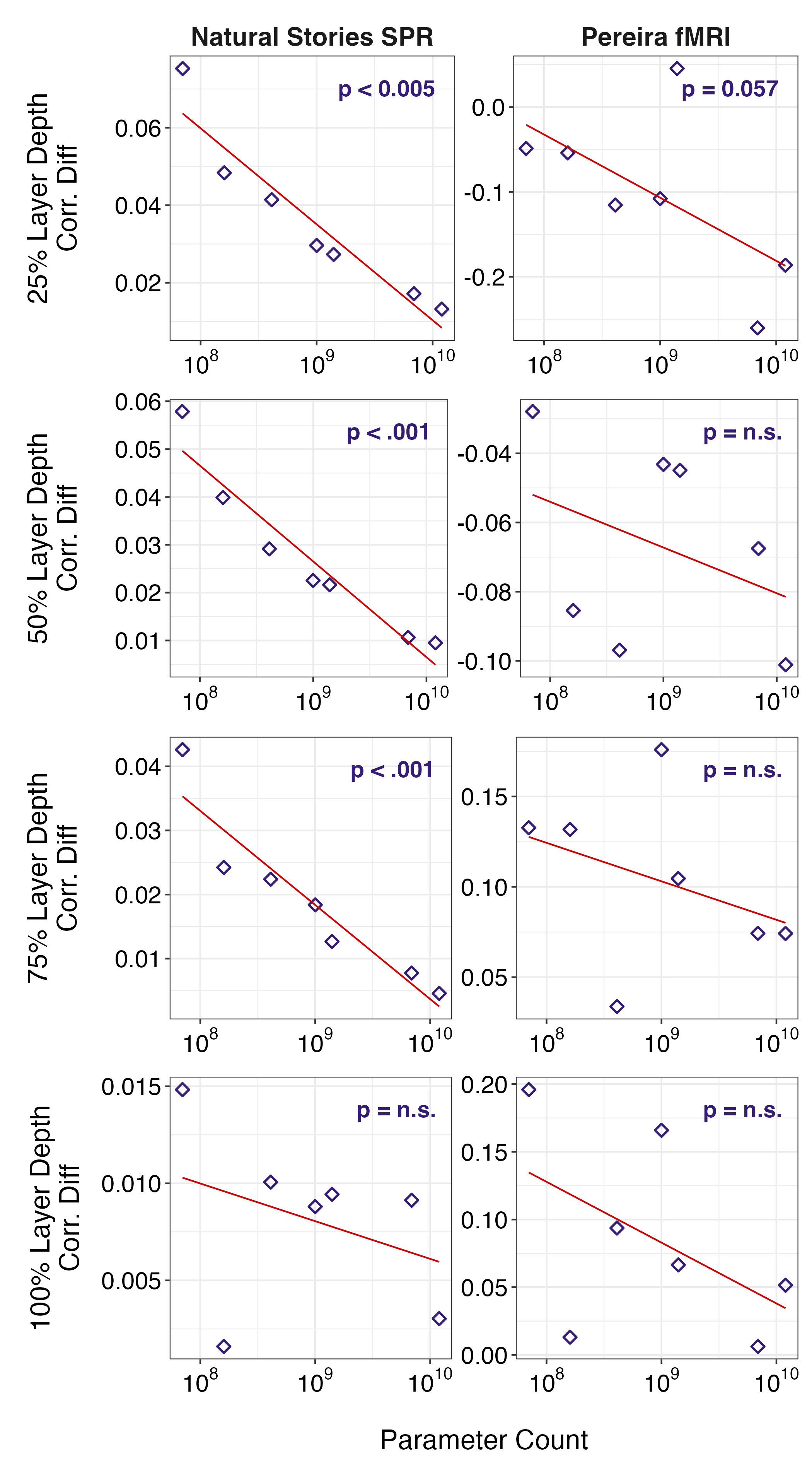}
    \caption{Normalized contribution (measured using correlation scores from linear regression models) full training to the predictive power of vector elements from fully trained Pythia models beyond the effect of degrees of freedom or RC.}
    \label{fig:linear-reg-norm-corr-subtraction}
\end{figure}

\begin{figure}[t!]
    \centering
    \includegraphics[width=0.85\columnwidth]{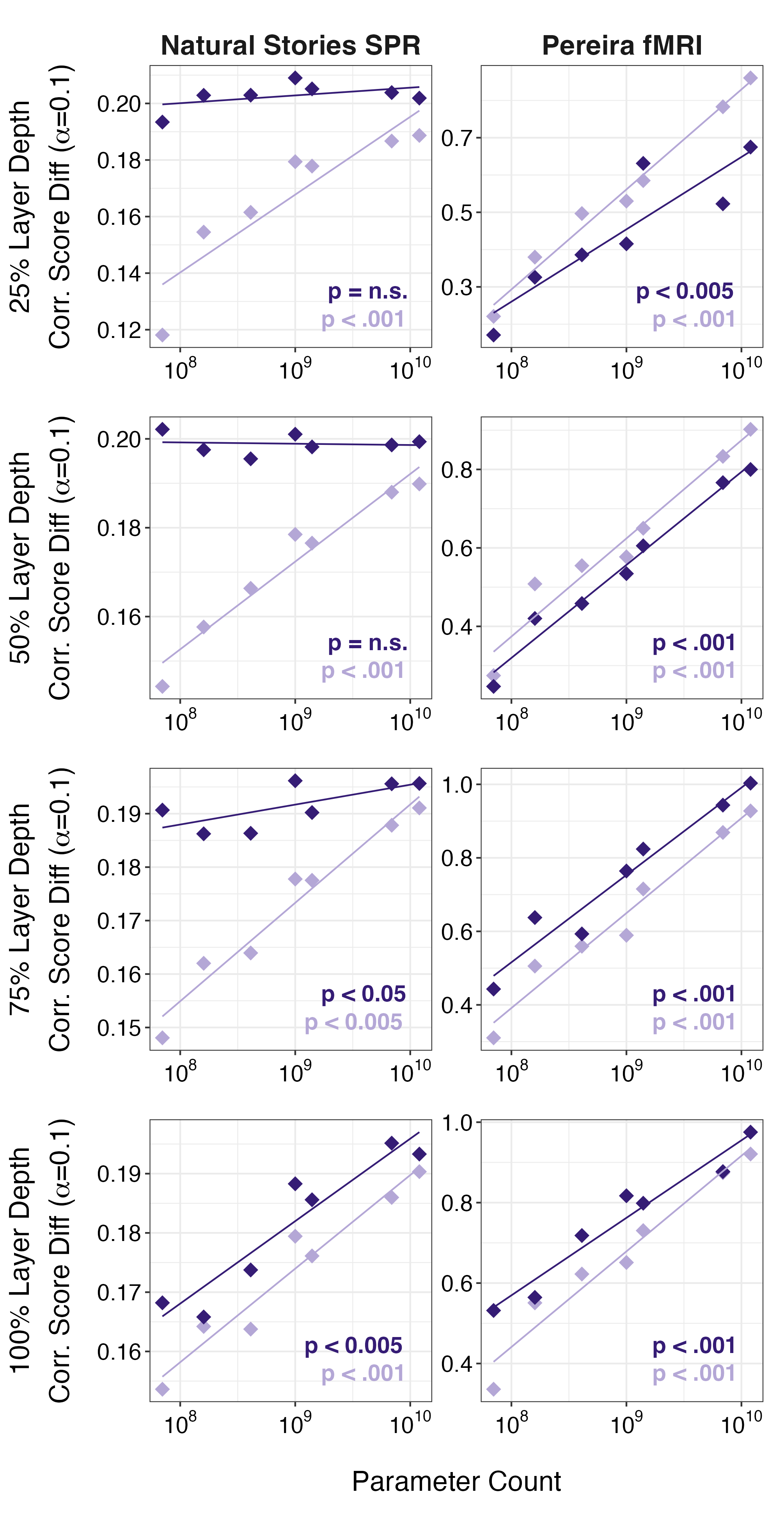}
    \caption{Predictive power (Pearson's correlation scores) of untrained (indicated with light purple) and fully-trained (indicated with dark purple) Pythia model vectors from different layer depths. The correlation scores reported in this figure are from ridge regression models ($\alpha$~=~0.1) and are normalized with the ceiling values provided in \citet{schrimpf2021neural}.}
    \label{fig:ridge-reg-alpha0.1-normalized-corr-without-baseline-sub}
\end{figure}

\begin{figure}[t!]
    \centering
    \includegraphics[width=0.9\columnwidth]{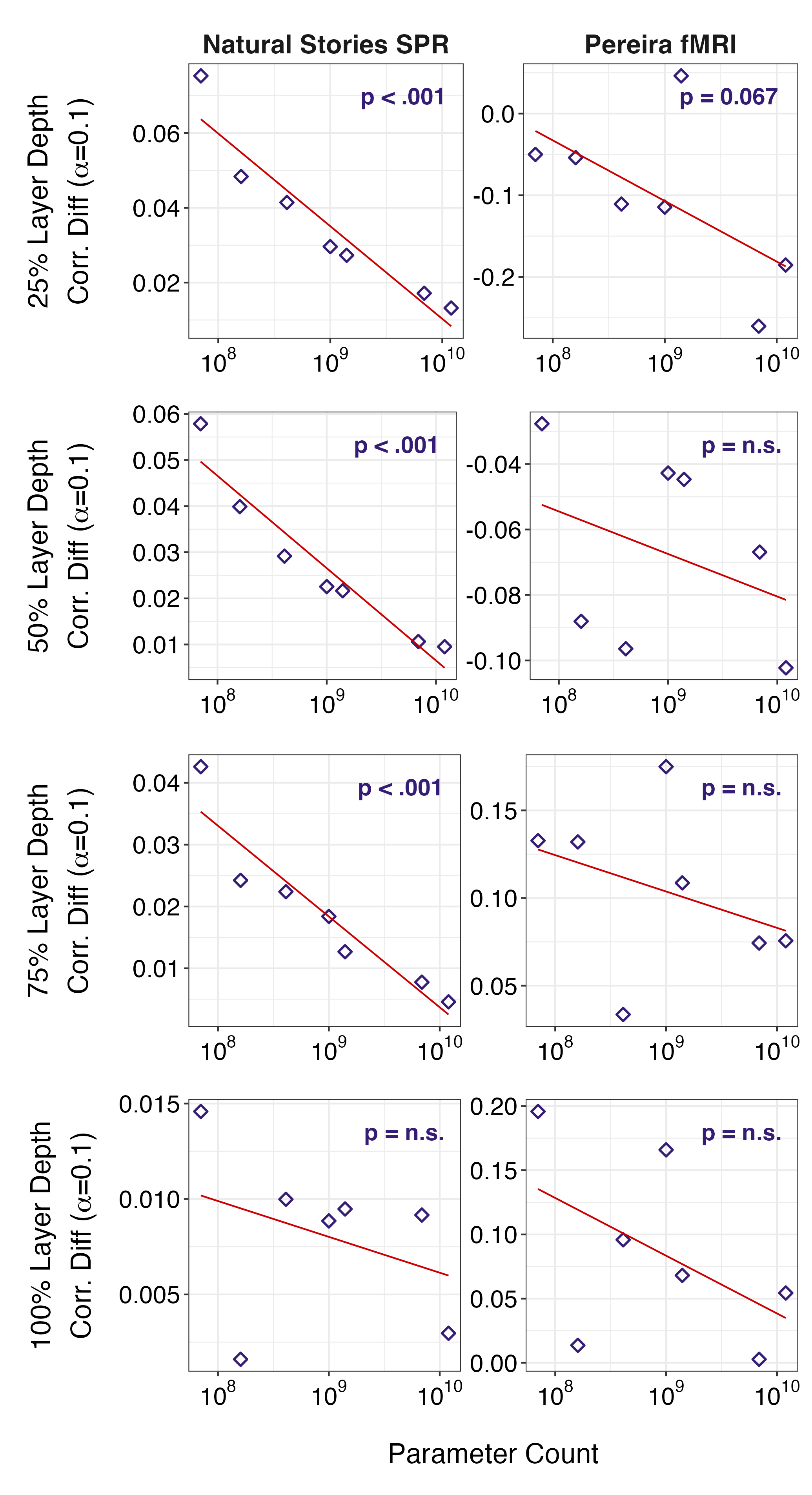}
    \caption{Normalized contribution (measured using correlation scores from ridge regression models ($\alpha$~=~0.1)) full training to the predictive power of vector elements from fully trained Pythia models beyond the effect of degrees of freedom or RC.}
    \label{fig:ridge-reg-alpha0.1-norm-corr-subtraction}
\end{figure}

\begin{figure}[t!]
    \centering
    \includegraphics[width=0.85\columnwidth]{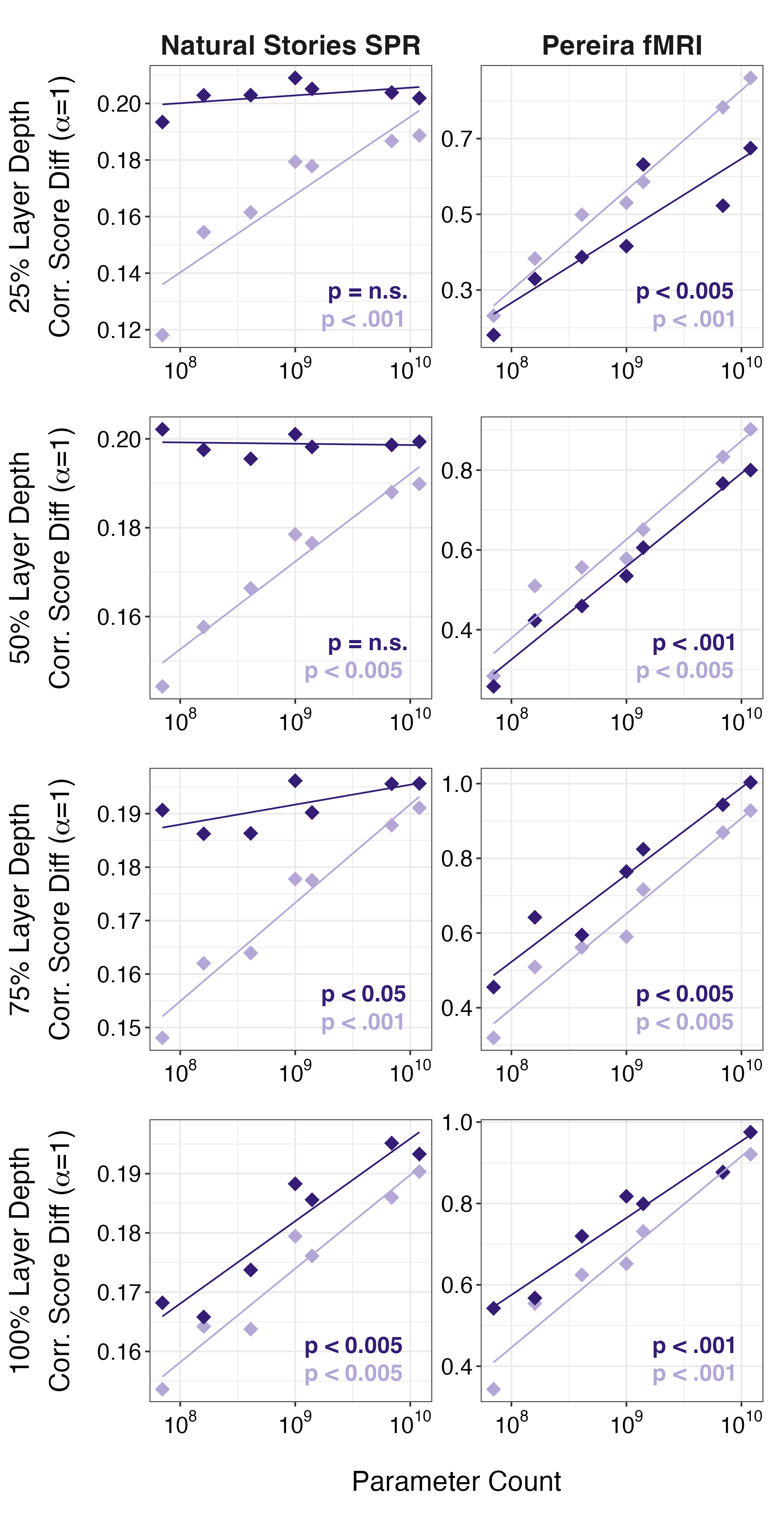}
    \caption{Predictive power (Pearson's correlation scores) of untrained (indicated with light purple) and fully-trained (indicated with dark purple) Pythia model vectors from different layer depths. The correlation scores reported in this figure are from ridge regression models ($\alpha$~=~1) and are normalized with the ceiling values provided in \citet{schrimpf2021neural}.}
    \label{fig:ridge-reg-alpha1-normalized-corr-without-baseline-sub}
\end{figure}

\begin{figure}[t!]
    \centering
    \includegraphics[width=0.9\columnwidth]{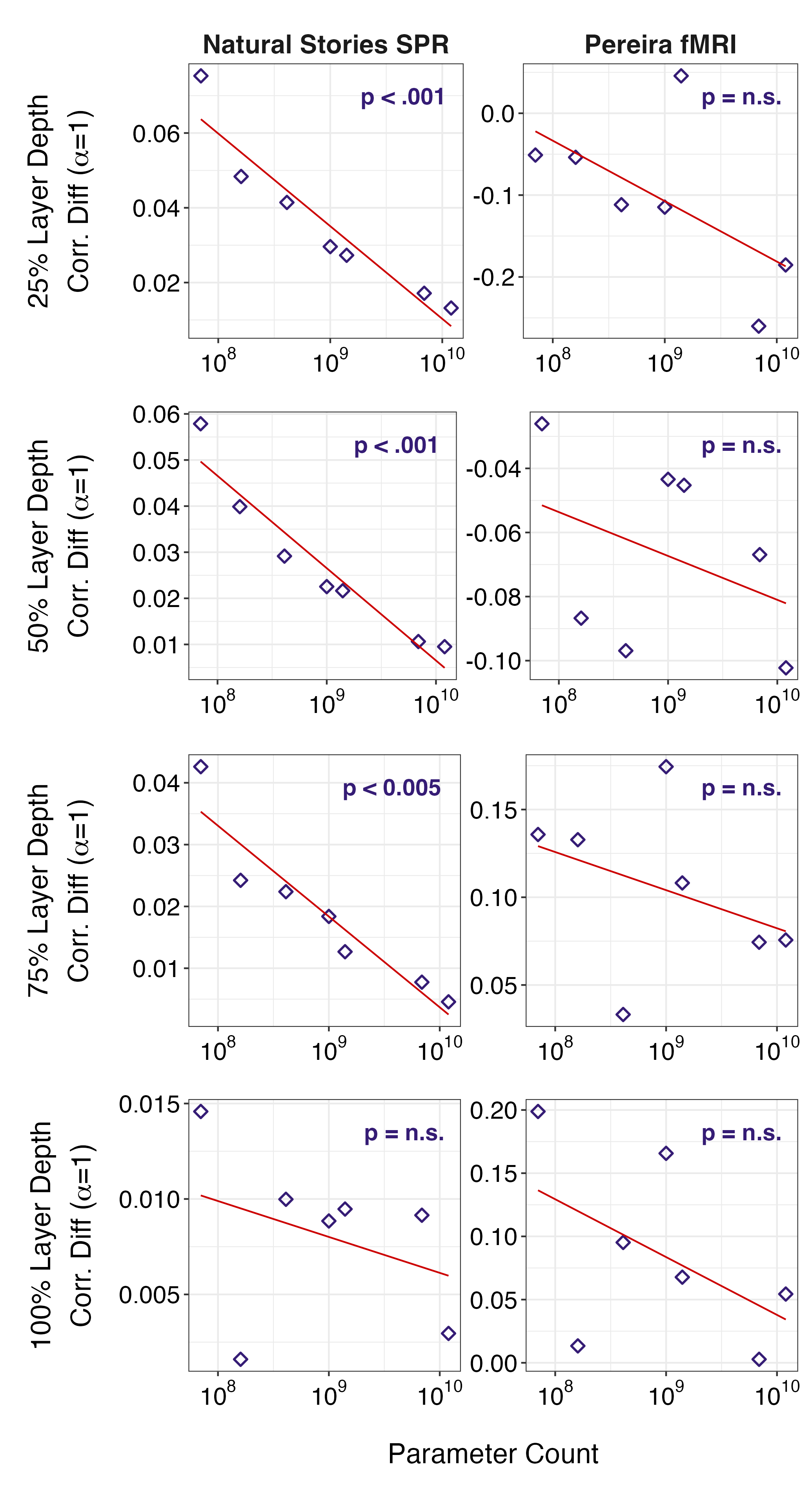}
    \caption{Normalized contribution (measured using correlation scores from ridge regression models ($\alpha$~=~1)) full training to the predictive power of vector elements from fully trained Pythia models beyond the effect of degrees of freedom or RC.}
    \label{fig:ridge-reg-alpha1-norm-corr-subtraction}
\end{figure}

For both linear (Figure~\ref{fig:linear-reg-normalized-corr-without-baseline-sub}) and ridge (Figures~\ref{fig:ridge-reg-alpha0.1-normalized-corr-without-baseline-sub} and \ref{fig:ridge-reg-alpha1-normalized-corr-without-baseline-sub}) regression, after normalization, the predictive power of vectors from untrained and trained Pythia models shows patterns consistent with the results shown in Section~\ref{exp2}.
The predictive power of both untrained and trained Pythia models show a significant positive scaling for most layer depths for both datasets.
The contribution of full training beyond the effect of RC after normalization also shows inverse scaling consistent with results shown in Section~\ref{contribution}, regardless of the regression type (Figures~\ref{fig:linear-reg-norm-corr-subtraction}, \ref{fig:ridge-reg-alpha0.1-norm-corr-subtraction}, and \ref{fig:ridge-reg-alpha1-norm-corr-subtraction}).

For Natural Stories SPR, as shown in the left panels of Figures~\ref{fig:linear-reg-normalized-corr-without-baseline-sub}, \ref{fig:ridge-reg-alpha0.1-normalized-corr-without-baseline-sub}, and \ref{fig:ridge-reg-alpha1-normalized-corr-without-baseline-sub}, after normalization, the predictive power of untrained and trained Pythia model vectors is still far from the upper bound for Natural Stories SPR (i.e., the normalized correlation scores for all untrained and trained LMs are far below 1).
Therefore, the inverse scaling of the contribution of training is not due to the confound of the ceiling effect.

On the other hand, the normalized predictive power of Pereira fMRI shifts closer to the upper bound (right panels of Figures~\ref{fig:linear-reg-normalized-corr-without-baseline-sub}, \ref{fig:ridge-reg-alpha0.1-normalized-corr-without-baseline-sub}, and \ref{fig:ridge-reg-alpha1-normalized-corr-without-baseline-sub}).
However, only the predictive power of vectors from fully trained Pythia 12B at the 75\textsuperscript{th} of layer depth reaches the upper bound (i.e., having a normalized correlation score greater than or equal to 1).
Other than this case, the normalized predictive power of vectors at layer depths from untrained and trained Pythia models does not reach the upper bound.
Since the contribution of training drops to zero even for smaller models and the normalized power of these models is far from the upper bound, a ceiling effect seems unlikely to be a confound for the zero contribution results of Pereira fMRI.

\end{document}